\documentclass[twoside]{article}
\usepackage[accepted]{aistats2021}

\usepackage[round]{natbib}

\usepackage[utf8]{inputenc} 
\usepackage[T1]{fontenc}    
\usepackage{hyperref}       
\usepackage{url}            
\usepackage{booktabs}       
\usepackage{amsfonts}       
\usepackage{nicefrac}       
\usepackage{microtype}      
\usepackage{wrapfig}
\usepackage{breqn}
\usepackage{hyperref}
\usepackage{algorithm}
\usepackage{algorithmic}
\usepackage{amsmath,amsthm,amssymb,comment}
\usepackage{pgf, tikz}
\usepackage{graphicx,float}
\usetikzlibrary{arrows,shapes,positioning}
\usetikzlibrary{calc,decorations.markings,automata}

\usepackage{soul}

\newcommand{\ATT}{\mbox{\tiny{ATT}}}
\newcommand{\ATE}{\mbox{\tiny{ATE}}}
\newcommand{\PEHE}{\mbox{\tiny{PEHE}}}
\newcommand{\EB}{\mbox{\tiny{EB}}}

\newcommand{\logit}{\mbox{logit}}
\newcommand{\argmax}{\mbox{arg~max}}
\newcommand{\ind}{\perp \! \! \! \perp}

\newcommand{\BR}{\mathbb{R}}
\newcommand{\BD}{\mathbb{D}}
\newcommand{\BH}{\mathbb{H}}
\newcommand{\IPM}{\textup{IPM}}
\newcommand{\KL}{\textup{KL}}
\newcommand{\JSD}{\textup{JSD}}

\usepackage[textwidth=1.2in,textsize=tiny]{todonotes}

\newtheorem{definition}{{\sc Definition}}
\newtheorem{assumption}{{\sc Assumption}} 
\newtheorem{theorem}{{\sc Theorem}}
\newtheorem{proposition}{{\sc Proposition}}

\usepackage{amsmath}
\usepackage{amsfonts}
\usepackage{amssymb}
\usepackage{wrapfig}
\usepackage{subcaption}
\usepackage{multirow}
 \usepackage{mathtools} 

\usepackage{verbatim}

\usepackage{anyfontsize}

\usepackage{color}
\usepackage{tikz}
\usetikzlibrary{arrows,shapes,snakes,automata,backgrounds,fit,petri}
\usepackage{adjustbox}

\makeatletter
\newcommand{\distas}[1]{\mathbin{\overset{#1}{\kern\z@\sim}}}%

\usepackage{enumitem}

\def\KL{\textsf{KL}} 

\newcommand{\beq}{\vspace{0mm}\begin{equation}}
\newcommand{\eeq}{\vspace{0mm}\end{equation}}
\newcommand{\beqs}{\vspace{0mm}\begin{eqnarray}}
\newcommand{\eeqs}{\vspace{0mm}\end{eqnarray}}
\newcommand{\barr}{\begin{array}}
\newcommand{\earr}{\end{array}}

\newcommand{\wv}{\boldsymbol{w}}

\newcommand{\lambdav}[0]{{\boldsymbol{\lambda}}}

\ifx\theorem\undefined
\newtheorem{theorem}{Theorem} 
\fi

\ifx\lemma\undefined
\newtheorem{lemma}{Lemma}
\fi

\ifx\proposition\undefined
\newtheorem{proposition}[theorem]{Proposition}
\fi

\ifx\corollary\undefined
\newtheorem{corollary}{Corollary}
\fi

\ifx\assumption\undefined
\newtheorem{assumption}{Assumption}
\fi

\ifx\definition\undefined
\newtheorem{definition}{Definition}
\fi

\ifx\remark\undefined
\newtheorem{remark}{Remark}
\fi

\begin{document}

%

%

\twocolumn[

\aistatstitle{Double Robust Representation Learning for Counterfactual Prediction}

\aistatsauthor{Shuxi Zeng \And Serge Assaad \And Chenyang Tao \And Shounak Datta \And Lawrence Carin \And Fan Li }

\aistatsaddress{ Duke University } ]

\begin{abstract}
Causal inference, or counterfactual prediction, is central to decision making in healthcare, policy and social sciences. To de-bias causal estimators with high-dimensional data in observational studies, recent advances suggest the importance of combining machine learning models for both the propensity score and the outcome function. We propose a novel scalable method to learn double-robust representations for counterfactual predictions, leading to consistent causal estimation if the model for either the propensity score or the outcome, but not necessarily both, is correctly specified. Specifically, we use the entropy balancing method to learn the weights that minimize the Jensen-Shannon divergence of the representation between the treated and control groups, based on which we make robust and efficient counterfactual predictions for both individual and average treatment effects. We provide theoretical justifications for the proposed method. The algorithm shows competitive performance with the state-of-the-art on real world and synthetic data. 
\end{abstract}

\section{INTRODUCTION}
Causal inference is central to decision-making in healthcare, policy, online advertising and social sciences. The main hurdle to causal inference is confounding, $i.e.$, factors that affect both the outcome and the treatment assignment \citep{vanderweele2013definition}. For example, a beneficial medical treatment may be more likely assigned to patients with worse health conditions; then directly comparing the clinical outcomes of the treated and control groups, without adjusting for the difference in the baseline characteristics, would severely bias the causal comparisons and mistakenly conclude the treatment is harmful. Therefore, a key in de-biasing causal estimators is to balance the confounding covariates or features. 

This paper focuses on using observational data to estimate treatment effects, defined as the contrasts between the counterfactual outcomes of the same study units under different treatment conditions \citep{Neyman1923, rubin1974estimating}. In observational studies, researchers do not have direct knowledge on how the treatment is assigned, and substantial imbalance in covariates between different treatment groups is prevalent. A classic approach for balancing covariates is to assign an importance weight to each unit so that the covariates are balanced after reweighting \citep{hirano2003WEIGHTING,hainmueller2012entropy,imai2014cbps,li2018balancing,kallus2018balanced}. The weights usually involve the \textit{propensity score} \citep{rosenbaum1983a} -- a summary of the treatment assignment mechanism. Another stream of conventional causal methods directly model the outcome surface as a function of the covariates under treated and control condition to impute the missing counterfactual outcomes \citep{Rubin79,imbens2005regress,hill2011bart}. 

Advances in machine learning bring new tools to causal reasoning. A popular direction employs the framework of representation learning and impose balance in the representation space \citep{johansson2016learning,shalit2017estimating,zhang2020learning}. These methods usually separate the tasks of propensity score estimation and outcome modeling. However, recent theoretical evidence reveals that good performance in predicting either the propensity score or the observed outcome alone does not necessarily translate into good performance in estimating the causal effects \citep{belloni2014highdimensionalinference}. In particular, \citep{chernozhukov2018double} pointed out it is necessary to combine machine learning models for the propensity score and the outcome function to achieve $\sqrt{N}$ consistency in estimating the average treatment effect (ATE). A closely related concept is double-robustness \citep{scharfstein1999, Lunceford04, kang2007demystifying}, in which an estimator is consistent if either the propensity score model or the outcome model, but not necessarily both, is correctly specified. A similar concept also appears in the field of reinforcement learning for policy evaluation \citep{dudik2011doublyRL,jiang2016doubly,kallus2019double}. Double-robust estimators are desirable because they give analysts two chances to ``get it right'' and guard against model misspecification.

In this work, we bridge the classical and modern views of covariate balancing in causal inference in a unified framework. We propose a novel and scalable method to learn double-robust representations for counterfactual predictions with observational data, allowing for robust learning of the representations and balancing weights simultaneously. Though the proposed method is motivated by ATE estimation, it also achieves comparable performance with state-of-the-art on individual treatment effects (ITE) estimation. Specifically, we 
made the following contributions: (i) We propose to regularize the representations with the entropy of an optimal weight for each unit, obtained via an entropy balancing procedure. (ii) We show that minimizing the entropy of balancing weights corresponds to a regularization on Jensen-Shannon divergence of the low-dimensional representation distributions between the treated and control groups, and more importantly, leads to a double-robust estimator of the ATE. (iii) We show that the entropy of balancing weights can bound the generalization error and therefore reduce ITE prediction error.
\vspace{-1.5em}
\section{BACKGROUND}
\vspace{-0.5em}
\label{sec:background}
\subsection{Setup and Assumptions}
\vspace{-0.5em}
\label{sec:setup}
Assume we have a sample of $N$ units, with $N_0$ in treatment group and $N_1$ in control group. Each unit $i$ ($i=1,2,\cdots,N$) has a binary treatment indicator $T_{i}$ ($T_{i}=0$ for control and $T_{i}=1$ for treated), $p$ features or covariates $\mathbf{X}_{i}=(X_{1i},\cdots,X_{ji},\cdots, X_{pi})\in \mathcal{R}^{p}$. Each unit has a pair of potential outcomes $\{Y_{i}(1),Y_{i}(0)\}$ corresponding to treatment and control, respectively, and causal effects are contrasts of the potential outcomes. We define individual treatment effect (ITE), also known as conditional average treatment effect (CATE) for context $x$ as: $\tau(x)=E\{Y_{i}(1)-Y_{i}(0)|X_{i}=x\}$, and the average treatment effect (ATE) as: $\tau_{\ATE}=E\{Y_{i}(1)-Y_{i}(0)\}=E_x\{\tau(x)\}$.
The ITE quantifies the effect of the treatment for the unit(s) with a specific feature value, whereas ATE quantities the average effect over a target population. When the treatment effects are heterogeneous, the discrepancy between ITE for some context and ATE can be large. Despite the increasing attention on ITE in recent years, average estimands such as ATE remain the most important and commonly reported causal parameters in a wide range of disciplines. Our method is targeted at estimating ATE, but we will also examine its performance in estimating ITE.   

For each unit, only the potential outcome corresponding to the observed treatment condition is observed, $Y_{i}=Y_{i}(T_i)=T_{i}Y_{i}(1)+(1-T_{i})Y_{i}(0)$, and the other is counterfactual. Therefore, additional assumptions are necessary for estimating the causal effects. Throughout the discussion, we maintain two standard assumptions:
\begin{assumption}[Ignorabililty]
\label{A.1}
\hspace{-0.5em}$\{Y_{i}(1),Y_{i}(0)\} \ind T_{i} \mid X_{i}$
\end{assumption}
\begin{assumption}[Overlap]
\label{A.2}\hspace{-0.5em}
$0<P(T_{i}=1|X_{i})<1$.
\end{assumption}
\vspace{-0.5em}
Under Assumption \ref{A.1} and \ref{A.2}, treatment effects can be identified from the observed data. In observational studies, there is often significant imbalance in the covariates distributions between the treated and control groups, and thus directly comparing the average outcome between the groups may be lead to biased causal estimates. Therefore, an important step to de-bias the causal estimators is to balance the covariates distributions between the groups, which usually involves the propensity score $e(x)=P(T_{i}=1|X_{i}=x)$, a summary of the treatment assignment mechanism. Once good balance is obtained, one can also build an outcome regression model $f_t(x)=E(Y(t)|X_{i}=x)$ for $t=0,1$ to impute the counterfactual outcomes and estimate ATE and ITE via the vanilla estimator
$\hat{\tau}_{\ATE}=\sum_{i=1}^{N}\{\hat{f}_1(X_{i})-\hat{f}_0(X_{i})\}/N$ and $\hat{\tau}(x)=\hat{f}_1(x)-\hat{f}_0(x)$.
\vspace{-0.5em}
\subsection{Related Work}
\vspace{-0.5em}
\textbf{Double robustness} \quad
A \emph{double-robust} (DR) estimator combines the propensity score and outcome model; a common example for ATE \citep{robins1994estimation,Lunceford04} is:
\begin{small}
\vspace{-1em}
 \begin{equation}
 \label{eq:DR_general_form}
\begin{aligned}
\hat{\tau}_{\ATE}^{\textup{DR}}=\sum_{i=1}^{N}\hat{w}_{i}^{\textup{IPW}}(2T_{i}-1)\{Y_{i}-\hat{f}_{T_{i}}(X_{i})\}+\\
\frac{1}{N}\sum_{i=1}^{N}\{\hat{f}_{1}(X_{i})-\hat{f}_{0}(X_{i})\},
\end{aligned}
\end{equation}
\end{small}
\hspace{-0.3em}where $w_{i}^{\textup{IPW}}=\frac{T_{i}}{e(X_{i})}+\frac{(1-T_{i})}{1-e(X_{i})}$ is the inverse probability weights (IPW). DR estimator has two appealing benefits: (i) it is DR in the sense that it remains consistent if either propensity score model or outcome model is correctly specified, not necessarily both; (ii) it reaches the semiparametric efficiency bound of $\tau_{\ATE}$ if both models are correctly specified \citep{hahn1998, chernozhukov2018double}. However, the finite-sample variance for $\hat{\tau}_{\ATE}^{\textup{DR}}$ can be quite large when the IPW have extreme values, which is likely to happen with severe confoundings. Several variants of the DR estimator have been proposed to avoid extreme importance weights, such as clipping or truncation \citep{bottou2013counterfactual,wang2017optimal,su2019doubly}. We propose a new weighting scheme, combined with the representation learning, to calculate the weights with less extreme values and maintain the double robustness.
 
\textbf{Representation learning with balance regularization}\quad
For causal inference with high-dimensional or complex observational data, an important consideration is dimension reduction. Specifically, we may wish to find a representations $\Phi(\cdot)=[\Phi_{1}(\cdot),\Phi_{2}(\cdot),\cdots,\Phi_{m}(\cdot)]: \mathbb{R}^{p}\rightarrow \mathbb{R}^{m}$ of the original space, and build the model based on the representations $\Phi(x)$ instead of directly on the features $x$, $f_t(\Phi(x))$. To this end,  \cite{johansson2016learning} and \cite{shalit2017estimating} proposed to combine predictive power and covariate balance to learn the representations, via minimizing the following type of loss function in the Counterfactual Regression (CFR) framework:
\begin{small}
\begin{equation}
\label{eq:OF_general}
\begin{aligned}
\mbox{arg} \min_{f,\Phi}& \{\sum_{i=1} L(f_{T_{i}}(\Phi(\mathbf{X}_{i})),Y_{i})+\\
& \kappa \cdot\BD(\{\Phi(X_{i})\}_{T_{i}=0},\{\Phi(X_{i})\}_{T_{i}=1})\},
\end{aligned}
\end{equation}
\end{small}
\hspace{-0.5em}where the first term measures the predictive power the representation $\Phi$, the second term measures the distance between the representation distribution in treated and control groups, and $\kappa$ is a hyperparameter controlling the importance of distance. This type of loss function targets learning representations that are predictive of the outcome and well balanced between the groups. Choice of the distance measure $\BD$ in \eqref{eq:OF_general} is crucial for the operating characteristics of the method; popular choices include the Integral Probability Measure (IPM) such as the Wasserstein (WASS) distance \citep{villani2008optimal,cuturi2014fast} or Maximum Mean Discrepancy (MMD)\citep{gretton2009covariate}. 

Concerning related modifications of \eqref{eq:OF_general}, in \cite{zhang2020learning}, the authors argue that balancing representations in \eqref{eq:OF_general} may over-penalize the model when domain overlap is satisfied and propose to use the counterfactual variance as a measure for imbalance, which can also address measure the ``local'' similarity in distribution. In \cite{hassanpour2019counterfactual} the authors reweight regression terms with inverse probability weights (IPW) estimated from the representations. In \cite{johansson2018learning}, the authors tackle the distributional shift problem, for which they alternately optimize a weighting function and outcome models for prediction jointly to reduce the generalization error.

The optimization problem \eqref{eq:OF_general} only involves the outcome model $f_t(x)$, misspecification of which would likely introduce biased causal estimates. In contrast, the class of causal estimators of DR estimators like \eqref{eq:DR_general_form} combine the propensity score model with the outcome model to add robustness against model misspecifications. 
A number of DR causal estimators for high-dimensional data have been proposed \citep{belloni2014highdimensionalinference,farrell2015robust,antonelli2018doubly}, but none has incorporated representation learning. Below we propose the first DR representation learning method for counterfactual prediction. The key is the entropy balancing procedure, which we briefly review below.

\textbf{Entropy balancing} \quad
\label{sec:entropy_balancing}
To mitigate the extreme weights problem of IPW in \eqref{eq:DR_general_form}, one stream of weighting methods learn the weights by minimizing the variation of weights subject to a set of balancing constraints, bypassing estimating the propensity score. Among these, entropy balancing (EB) \citep{hainmueller2012entropy} has received much interest in social science \citep{ferwerda2014electoral,marcus2013effect}. EB was originally designed for estimating the average treatment effect on the treated (ATT), but is straightforward to adapt to other estimands. Specifically, the EB weights for ATE, are obtained via the following programming problem: 
\begin{small}
\begin{gather}
\label{eq:entropy_balancing}
\wv^{\EB} = \argmax_{w} \left\{-\sum_{i=1}^{N}w_{i}\log w_{i},\right\}, \,\,\\
\textrm{s.t. } \left\{ \begin{array}{l}
\textup{(i)} \sum_{T_{i}=0}w_{i} X_{ji}=\sum_{T_{i}=1}w_{i} X_{ji},  \forall j \in[1:p],\\
[5pt]
\textup{(ii)}\sum_{T_{i}=0}w_{i}=\sum_{T_{i}=1}w_{i}=1, w_{i}>0. 
\end{array}
\right.\notag
\end{gather}
\end{small}
\hspace{-0.8em} Covariate balancing is enforced by the the first constraint (i), also known as the moment constraint, that the weighted average for each covariate of respective treatment groups are equal. Generalizations to higher moments are straight forward although less considered in practice. The second constraint simply ensures the weights are normalized. This objective is an instantiation of the maximal-entropy learning principle \citep{jaynes1957information, jaynes1957informationii}, a concept derived from statistical physics that stipulates the most plausible state of a constrained physical system is the one maximizes its entropy. Intuitively, EB weights penalizes the extreme weights while keeps balancing condition satisfied. 

Though the construction of EB does not explicitly impose models for either $e(x)$ or $f_t(x)$, \cite{zhao2017entropy} showed that EB implicitly fits a linear logistic regression model for the propensity score and a linear regression model for the outcome simultaneously, where the predictors are the covariates pr representations being balanced. Entropy balancing is DR in the sense that if only of the two models are correctly specified, the EB weighting estimator is consistent for the ATE. Note that the original EB procedure does not provide ITE estimation, which is explored in this work.

\section{DOUBLE ROBUST REPRESENTATION LEARNING}
\label{sec:method}
\subsection{Proposal: Unifying Covariate Balance and Representation Learning}
\label{sec:proposal}
\begin{figure}[ht]
\vspace{-1em}
\begin{center}
\centerline{\includegraphics[width=1.2\linewidth]{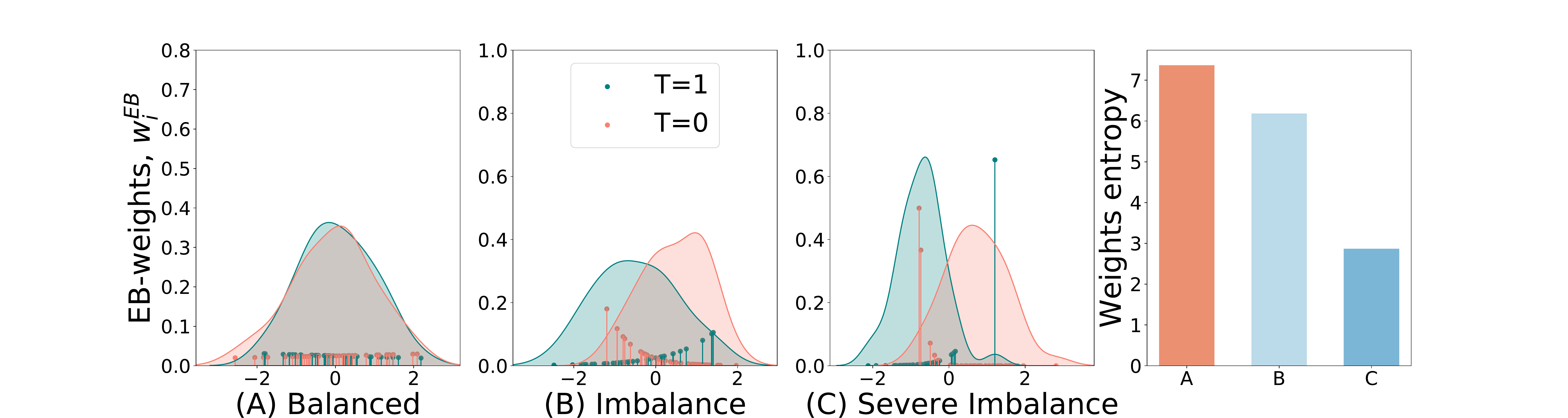}}
\caption{When covariates imbalance is more severe, the balance weights $w_{i}^{\EB}$ deviate more from uniform distribution, inducing a lower entropy}
\label{fig:illu_of_EB}
\end{center}
\end{figure}

Based on the discussion above, we propose a novel method to learn DR representations for counterfactual predictions. Our development is motivated by an insightful heuristic: the entropy of balancing weight is a proxy measure of the covariate imbalance between the treatment groups. To understand the logic behind this intuition, recall the more dis-similar two distributions are, the more likely extreme weights are required to satisfy the matching criteria, and consequently resulting a bigger entropy for the balancing weight. See Figure \ref{fig:illu_of_EB} also for a graphical illustration of this. In Section \ref{sec:theorem}, we will formalize this intuition based on information-theoretic arguments. 
\vspace{-2em}

We adjust the constrained EB programming problem from \eqref{eq:entropy_balancing} to \eqref{eq:entropy_balancing_representation}, achieving the balance among the representations/transformed features. As we shall see later, this distance metric, entropy of balancing weights, leads to desirable theoretical properties in both ATE and ITE estimation.
\begin{small}
\begin{gather}
\label{eq:entropy_balancing_representation}
\wv^{\EB} = \argmax_{w} \left\{-\sum_{i=1}^{N}w_{i}\log w_{i},\right\}, \,\,\\
\textrm{s.t. } \left\{ \begin{array}{l}
\textup{(i)} \sum_{T_{i}=0}w_{i} \Phi(X_{ji})=\sum_{T_{i}=1}w_{i} \Phi(X_{ji}),  \\
[5pt]
\textup{(ii)}\sum_{T_{i}=0}w_{i}=\sum_{T_{i}=1}w_{i}=1, w_{i}>0. 
\end{array}\notag
\right.
\end{gather}
\end{small}
\hspace{-0.4em}Specifically, we propose to learn a low-dimensional representation of the feature space, $\Phi(\cdot)$, through minimizing the following loss function:
\begin{small}
\begin{eqnarray}
\begin{aligned}
\label{eq:OF_EB}
\mbox{arg} \min_{f,\Phi} \overbrace{\{ \sum_{i}(Y_{i}-f_{t=T_i}(\Phi(X_{i})))^{2}}^{\textup{prediction loss on observed outcomes}}+\\
\underbrace{\kappa \sum_{i=1} w_{i}^{\EB}(\Phi)\log w_{i}^{\EB}(\Phi)\}}_{\textup{distance metric, balance regularization}},\hspace{-4em}
\end{aligned}
\end{eqnarray}
\end{small}
\hspace{-0.3em}where we replace the distance metrics in \eqref{eq:OF_general} with the entropy of $w_{i}^{\EB}(\phi)$, function of the representation as implied in the notation, which is the solution to \eqref{eq:entropy_balancing_representation}. At first sight, solving the system defined by \eqref{eq:entropy_balancing_representation}  and \eqref{eq:OF_EB} is challenging, because the gradient can not be back-propagated through the nested optimization \eqref{eq:entropy_balancing_representation}.
Another appealing property of EB is computational efficiency. We can solve the dual problem of \eqref{eq:entropy_balancing_representation}:
\begin{small}
\vspace{-1em}
\begin{equation}
\begin{aligned}
\label{eq:dual_entropy}
\min_{\lambdav}  &\{ \log \left(\sum_{T_{i}=0}\exp\left(\langle \lambdav_0, \Phi_i \rangle \right) \right) +\\ &\log \left(\sum_{T_{i}=1}\exp\left(\langle \lambdav_1, \Phi_i \rangle \right) \right) - \langle \lambdav_0 + \lambdav_1, \bar{\Phi} \rangle \},
\end{aligned}
\end{equation}
\end{small}
\hspace{-1pt}where $\lambdav_0, \lambdav_1 \in \BR^m$ are the Lagrangian multipliers, $\bar{\Phi} \triangleq \sum_{i}\Phi_i$ is the unnormalized mean and $\langle\cdot ,\cdot\rangle$ denote the inner product. Note that \eqref{eq:dual_entropy} is a convex problem wrt $\lambdav$, and therefore can be efficiently solved using standard convex optimization packages when the sample size is small.  
Via appealing to the Karush–Kuhn–Tucker (KKT) conditions, the optimal EB weights $\wv^{\EB}$ can be given in the following Softmax form
\begin{small}
\begin{equation}
\begin{aligned}
\label{eq:softmax}
w_{i}^{\EB}(\Phi)=\frac{\exp(\eta_i)}{\sum_{T_{k}=T_{i}}\exp(\eta_k)}, \\
\eta_i \triangleq -(2T_{i}-1) \langle \lambdav_{T_i}^{\EB}, \Phi_i \rangle ,
\end{aligned}
\end{equation}
\end{small}
\hspace{-0.3em}where $\lambdav_t^{\EB}, t \in \{ 0, 1\}$ is the solution to the dual problem \eqref{eq:dual_entropy}. Equation \eqref{eq:softmax} shows how to explicitly express the entropy weights as a function of the representation $\Phi$, thereby enabling efficient end-to-end training of the representation. Compared to the CFR framework, we have replaced the IPM matching term $\BD_{\IPM}(q_0 \parallel q_1)$ with the entropy term $\BH(\wv^{\EB}) = \sum_i w_i^{\EB} \log w_i^{\EB}$. When applied to the ATE estimation, the commensurate learned entropy balancing weights $\wv^{\EB}$ guarantees the $\tau_{\ATE}(\wv^{\EB})$ to be DR. For ITE estimation, $\BH(\wv^{\EB})$, as a regularization term in \eqref{eq:OF_EB}, can bound the ITE prediction error. 

A few remarks are in order. For reasons that will be clear in Section \ref{sec:theorem}, we will restrict $f_t$ to the family of linear functions, to ensure the nice theoretical properties of DRRL. Note that is not a restrictive assumption, as many schemes seek representations that can linearize the operations. For instance, outputs of a deep neural nets are typically given by a linear mapping of the penultimate layers. Many modern learning theories, such as reproducing kernel Hilbert space (RKHS), are formulated under inner product spaces ({\it i.e.}, generalized linear operations).

After obtaining the representation $\hat{\Phi}(x)$, the outcome function $\hat{f}_t$, and the EB weights $\hat{w}_{i}^{\EB}$, we have the following estimators of $\tau_{\ATE}$ and $\tau(x)$, 
\begin{small}
\vspace{-1em}
\begin{eqnarray}
\hat{\tau}_{\ATE}^{\EB}&=&\sum_{i=1}^{N}\hat{w}_{i}^{\EB}(2T_{i}-1)\{Y_{i}-\hat{f}_{T_{i}}(\hat{\Phi}(X_{i}))\}\nonumber\\
&&+\frac{1}{N}\sum_{i=1}^{N}\{\hat{f}_{1}(\hat{\Phi}(X_{i}))-\hat{f}_{0}(\hat{\Phi}(X_{i}))\},\label{eq:ATE_EB}\\
\hat{\tau}^{\EB}(x)&=&\hat{f}_{{1}}(\hat{\Phi}(x))-\hat{f}_{{0}}(\hat{\Phi}(x)). \label{eq:ITE_EB}
\end{eqnarray}
\end{small}
\hspace{-0.4em}In practice, we can parameterize the representations by $\theta$ as $\Phi_{\theta}(\cdot)$ and the outcome function by $\gamma=(\gamma_{0},\gamma_{1})$ as $f_{t,\gamma}(\cdot)=f_{\gamma_{t}}(\cdot)=\langle \gamma_{t},\Phi_{\theta}\rangle$ to learn the $\theta,\gamma$ instead. 

\subsection{Practical Implementation}

\label{sec:implement}
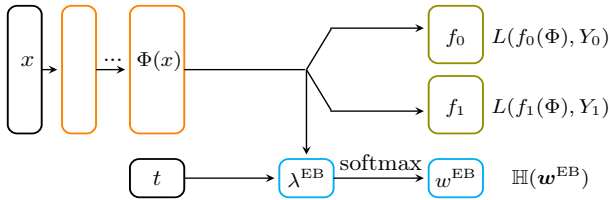
\begin{figure}[ht]
\centering
\resizebox{0.5\textwidth}{0.15\textwidth}{
\begin{tikzpicture}
[
> = stealth, 
shorten > = 2pt, 
auto,
node distance =4cm, 
semithick 
]
\tikzstyle{every state}=[
draw = black,
thick,shape=rectangle,
rounded corners,
fill = white,
minimum size = 1mm,
]
\draw[draw=black,rounded corners,thick] (0,0) rectangle node(A){} (0.5,2.0);
\node[align=center] at (A.center) {$\ \ \ x$}; 
\draw[draw=black,rounded corners,thick,orange] (0.8,0)rectangle node(B){ } (1.3,2.0);
\draw[draw=black,rounded corners,thick,orange] (1.8,0)rectangle node(C){} (2.6,2.0);
\node[align=center,font=\small] at (C.center) {$\ \ \ \Phi(x) $}; 
\draw[draw=black,rounded corners,thick,olive] (6.2,0)rectangle node(D){} (7.0,0.9); 
\node[align=center,font=\small] at (8.0,0.4) {$L(f_{1}(\Phi),Y_{1})$}; 
\node[align=center,font=\small] at (8.0,1.5) {$L(f_{0}(\Phi),Y_{0})$}; 
\node[align=center,font=\small] at (6.6,0.4) {$f_{1}$}; 
\node[align=center,font=\small] at (6.6,1.5) {$f_{0}$}; 

\draw[draw=black,rounded corners,thick,olive] (6.2,1.1)rectangle node(E){} (7.0,2.0);

\draw[draw=black,rounded corners,thick] (1.8,-1.0) rectangle node(F){} (2.6,-0.4);
\node[align=center] at (2.2,-0.7) {$t$}; 

\draw[draw=black,rounded corners,thick,cyan] (4.0,-1.0) rectangle node(G){} (4.8,-0.4);
\node[align=center,font=\small] at (4.4,-0.7) {$\lambda^{\EB}$}; 

\draw[draw=black,rounded corners,thick,cyan] (6.2,-1.0) rectangle node(H){} (7.0,-0.4);
\node[align=center,font=\small] at (6.6,-0.7) {$w^{\EB}$}; 
\node[align=center,font=\small] at (8.0,-0.7) {$\BH(\wv^{\EB})$};

\path[->](0.5,1.0) edge node {}(0.8,1.0);
\path[->](1.3,1.0) edge node {...}(1.8,1.0);
\path[-](2.6,1.0) edge node {}(4.45,1.0);
\path[-](4.4,1.0) edge node {}(4.8,0.3);
\path[-](4.4,1.0) edge node {}(4.8,1.7);
\path[->](4.75,1.65) edge node {}(6.2,1.65);
\path[->](4.75,0.35) edge node {}(6.2,0.35);

\path[->](4.4,1.0) edge node {}(4.4,-0.4);
\path[->](2.6,-0.7) edge node {}(4.0,-0.7);
\path[->](4.8,-0.7) edge node {softmax}(6.2,-0.7);
\end{tikzpicture}
}
\caption{Architecture of the DRRL network}
\label{network-archi}
\end{figure}

We now propose an algorithm -- referred as \textit{Double Robust Representations Learning} (DRRL) -- to implement the proposed method when we parameterize the representations $\Phi_{\theta}$ by neural networks. DRRL simultaneously learn the representations $\Phi_{\theta}$, the EB weights $w_{i}^{\EB}$ and the outcome function $f_{t,\gamma}$. The network consists of a representation layer performing non-linear transformation of the original feature space, an entropy-balancing layer solving the dual programming problem in \eqref{eq:dual_entropy} and a final layer learning the outcome function. We visualize the DRRL architecture in Figure \ref{network-archi}.

We train the model by iteratively solving the programming problem in \eqref{eq:entropy_balancing_representation} given the representations $\Phi$ and minimizing the loss function in \eqref{eq:OF_EB} given the optimized weights $w_{i}^{\EB}$. As we have successfully expressed EB weights, and consequently the entropy term, directly through the learned representation $\Phi$ in \eqref{eq:softmax}, it enables efficient gradient-based learned schemes, such as stochastic gradient descent, for the training of DRRL using modern differential programming platforms ({\it e.g.}, tensorflow, pytorch). 
As an additional remark, we note although the Lagrangian multiplier $\lambdav$ is computed from the representation $\Phi$, its gradient with respect to $\Phi$ is zero based on the Envelop theorem \citep{carter2001foundations}. This impliess we can safely treat $\lambdav$ as if it is a constant in our training objective. 


\begin{algorithm}[tb]
  \caption{Double Robust Representation Learning}
  \label{alg:drrl}
\begin{algorithmic}
  \STATE {\bfseries Input:} data $\{Y_{i},T_{i},X_{i}\}_{i=1}^{N}$, 
  \STATE {\bfseries Hyperparameters:} importance of balance $\kappa$, dimension of representations $m$, batch size $B$,  learning rate $\eta$.
  \STATE Initialize $\theta^{0},\gamma^{0},\lambdav^{0}$.
  \FOR{$k=1$ {\bfseries to} $K$}
  \STATE Sample batch data $\{Y_{i},X_{i},T_{i}\}_{i=1}^{B}$
  \STATE Calculate $\Phi(X_{i})=\Phi_{\theta^{k-1}}(X_{i})$ for each $i$ in the batch
    \STATE Entropy balance steps: Calculate the gradient of objective in \eqref{eq:dual_entropy} with respect to $\lambdav$, $\triangledown_{\lambdav}$, update $\lambdav^{k}=\lambdav^{k-1}-\eta\triangledown_{\lambdav}$.
  \STATE Learn representations and outcome function: calculate the gradient of loss \eqref{eq:OF_EB} in the batch data with respect to $\theta$ and $\gamma$, $\triangledown_{\theta},\triangledown_{\gamma}$. Update the parameters: $\theta^{k}=\theta^{k-1}-\eta\triangledown_{\theta}$,$\gamma^{k}=\gamma^{k-1}-\eta\triangledown_{\gamma}$.
  \ENDFOR
  \STATE Calculate the weights $w_{i}^{\EB}$ with formula \eqref{eq:softmax}.
  \STATE{\bfseries{Output} $\Phi_{\theta}(\cdot),f_{t,\gamma},w_{i}^{\EB}$}
\end{algorithmic}
\end{algorithm}

\textbf{Adaptation to ATT estimand} \quad So far we have focused on DR representations for ATE; the proposed method can be easily modified to other estimands. For example, for the average treatment effect on  the treated (ATT), we can modify the EB constraint to $\sum_{T_{i}=0}w_{i} \Phi_{ji}=\sum_{T_{i}=1}\Phi_{ji}/N_{1}$ and change the objective function to $-\sum_{T_{i}=0}w_{i}\log w_{i}$ in \eqref{eq:entropy_balancing_representation}.
For ATT, we only need to reweight the control group to match the distribution of the treated group, which remains the same. Thus we only impose balancing constraints on the weighted average of representations of the control units; the objective function only applies to the weights of the control units. In the SM, we also provide theoretical proofs for the double-robustness property of the ATT estimator.

\textbf{Scalable generalization} \quad
A bottleneck in scaling up our algorithm to large data is solving optimization problem \eqref{eq:dual_entropy} in the entropy balancing stage. Below we develop a scalable updating scheme with the idea of Fenchel mini-max learning in \citet{taofenchel}. Specifically, let $g(d)$ be a proper convex, lower-semicontinuous function; then its convex conjugate function $g^{\ast}(v)$ is defined as $g^{\ast}(v)=\sup_{d\in \mathcal{D}(g)}\{dv-g(d)\}$, where $\mathcal{D}(g)$ denotes the domain of the function $g$ \citep{hiriart2012fundamentals}; $g^{\ast}$ is also known as the Fenchel conjugate of $g$, which is again convex and lower-semicontinuous. The Fenchel conjugate pair $(g,g^{\ast})$ are dual to each other, in the sense that $g^{\ast\ast}=g$, $i.e.$, $g(v)=\sup_{d\in \mathcal{D}(g^{\ast} }\{dv-g^{\ast}(d)\}$. As a concrete example, $(-\log(d),-1-\log(-v))$ gives such a pair, which we exploit for our problem. Based on the Fenchel conjugacy, we can derive the mini-max training rule for the entropy-balancing objective in
\eqref{eq:dual_entropy}, for $t=0,1$:
\begin{small}
\begin{equation}
\label{eq:fenchel}
\begin{aligned}
\min_{\lambdav_{t}}\{ \max_{u_{t}} 
\{u_{t}-\exp(u_{t})\sum_{T_{i}=t}\exp\left(\langle \lambdav_t, \Phi_i \rangle 
\right) \}-\langle \lambdav_t, \Phi_i \rangle \}.
\end{aligned}
\end{equation}
\end{small}

\vspace{-2em}
\subsection{Theoretical Properties}
\label{sec:theorem}
In this section we establish the nice theoretical properties of the proposed DRRL framework. Limited by space, detailed technical derivations on Theorem \ref{thm:EB_KL_equiv}, \ref{thm:DR} and \ref{thm:bound} are deferred to the SM. 

Our first theorem shows that, the entropy of the EB weights as defined in \eqref{eq:OF_EB} asymptotically converges to a scaled $\alpha$-Jensen-Shannon divergence (JSD) of the representation distribution between the treatment groups. 

\begin{theorem}[EB entropy as JSD]
\label{thm:EB_KL_equiv}
The Shannon entropy of the EB weights defined in \eqref{eq:entropy_balancing_representation} converges in probability to the following $\alpha$-Jensen-Shannon divergence between the marginal representation distributions of the respective treatment groups:
\begin{small}
\begin{equation}\label{eq:EB_KL}
\begin{aligned}
\lim_{n \rightarrow \infty} &\BH_n^{\EB}(\Phi)\triangleq \sum_{i}w_{i}^{\EB}(\Phi)\log (w_{i}^{\EB}(\Phi))\\ 
\stackrel{p}{\longrightarrow} &c'\{\KL(p_{\Phi}^{1}(x)||p_{\Phi}(x))+\KL(p_{\Phi}^{0}(x)||p_{\Phi}(x))\}+c''\\
=&c' \JSD{\alpha}(p_{\Phi}^{1},p_{\Phi}^{0})+c''
\end{aligned}
\end{equation}
\end{small}
\hspace{-0.4em}where $c'>0, c''$ are non-zero constants,  $p_{\Phi}^{t}(x)=P(\Phi(\mathbf{X}_{i}=x)|T_{i}=t)$ is representation distribution in group $t$ ($t=0,1$), $p_{\Phi}(x)$ is the marginal density of the representations, $\alpha$ is the proportion of treated units $P(T_{i}=1)$ and $\KL(\cdot||\cdot)$ is the Kullback–Leibler (KL) divergence.
\end{theorem}
An important insight from Theorem \ref{thm:EB_KL_equiv} is that entropy of EB weights is an endogenous measure of representation imbalance, validating the insight in Sec \ref{sec:proposal} theoretically. This theorem bridges the classical weighting strategies with the modern representation learning perspectives for causal inference,  that representation learning and propensity score modeling are inherently connected and does not need to be modeled separately.


\begin{theorem}[Double Robustness]
\label{thm:DR} 
Under the Assumption \ref{A.1} and \ref{A.2}, the entropy balancing estimator $\hat{\tau}_{\ATE}^{\EB}$ is consistent for $\tau_{\ATE}$  if
either the true outcome models $f_t(x),t\in\{0,1\}$ or the true propensity score model $\logit\{e(x)\}$ is linear in representation $\Phi(x)$.
\end{theorem}

Theorem \ref{thm:DR} establishes the DR property of the EB estimator $\hat{\tau}^{\EB}$. Note that the double robustness property will not be compromised if we add regularization term in \eqref{eq:OF_EB}. Double robust setups require modeling both the outcome function and propensity score; in our formulation, the former is explicitly specified in the first component in \eqref{eq:OF_EB}, while the latter is implicitly specified via the EB constraints in \eqref{eq:entropy_balancing_representation}. By M-estimation theory \citep{stefanski2002calculus}, we can show that $\lambda^{\EB}$ in \eqref{eq:dual_entropy} converges to the maximum likelihood estimate $\lambda^{\ast}$ of a logistic propensity score model, which is equivalent to the solution of the following optimization problem,
\begin{small}
\begin{equation}
\label{eq:logistic_regression}
\begin{aligned}
\min_{\lambda} \quad & \sum_{i=1}^{N}\log (1+\exp(-(2T_{i}-1)\sum_{j=1}^{m}\lambda_{j}\Phi_{j}(X_{i}))).
\end{aligned}
\end{equation}
\end{small}
\hspace{-0.4em}Jointly these two components constructs the double robustness property of estimator $\hat{\tau}_{\ATE}^{\EB}$. The linearity restriction on $f_{t}$ is essential for double robustness, and may appear to be tight, but because the representations $\Phi(x)$ can be complex functions such as multi-layer neural networks (as in our implementation), both the outcome and propensity score models are flexible.

The third theorem shows that the objective function in \eqref{eq:OF_EB} is an upper bound of the loss for the ITE. Before proceeding to the third theorem, we define a few estimation loss functions: Let $L(y,y')$ be the loss function on predicting the outcome, $l_{f,\Phi}(x,t)$ denote the expected loss for a specific covariates-treatment pair $(x,t)$ given outcome function $f$ and representation, 
\begin{small}
\begin{eqnarray}
\label{eq:expect_loss}
l_{f,\Phi}(x,t)=\int_{y}L(Y(t),f_{t}(\Phi_{x}))P(Y(t)|x)dY(t).
\end{eqnarray}
\end{small}
\hspace{-0.4em}Suppose the covariates follow $\mathbf{X}_{i}\in\mathcal{X}$ and we denote the distributions in treated and control group with $p_{t}(x)=p(X_{i}=x|T_{i}=t),t=0,1$. For a given $f$ and $\Phi$, the expected factual loss over the distributions in the treated and control groups are,
\begin{small}
\begin{eqnarray}
\label{eq:f_cf_loss}
\varepsilon_{\textup{F}}^{t}(f,\Phi)&=&\int_{\mathcal{X}}l_{f,\phi}(x,t)p_{t}(x)dx, t=0,1,
\end{eqnarray}
\end{small}
\hspace{-0.4em}For the ITE estimation, we define the expected Precision in Estimation of Heterogeneous Effect (PEHE) \citep{hill2011bart}, 
\begin{small}
\begin{eqnarray}
\begin{aligned}
\label{eq:PEHE_formula}
\varepsilon_{\textup{PEHE}}(f,\Phi)\hspace{-0.3em}=\hspace{-0.35em}
\int_{\mathcal{X}}(f_{1}(\Phi(x))-f_{0}(\Phi(x))-\tau(x))^{2}p(x)dx.
\end{aligned}
\end{eqnarray}
\end{small}
\hspace{-0.4em}Assessing $\varepsilon_{\textup{PEHE}}(f,\Phi)$ from the observational data is infeasible, as the countefactual labels are absent, but we can calculate the factual loss $\varepsilon_{\textup{F}}^{t}$. The next theorem illustrates we can bound $\varepsilon_{\textup{PEHE}}$ with $\varepsilon_{\textup{F}}^{t}$ and the $\alpha$-JS divergence of $\Phi(x)$ between the treatment and control groups.

\begin{theorem}
\label{thm:bound}
Suppose $\mathcal{X}$ is a compact space and $\Phi(\cdot)$ is a continuous and invertible function. For a given $f,\Phi$, the expected loss for estimating the ITE, $\varepsilon_{\textup{PEHE}}$, is bounded by the sum of the prediction loss on the factual distribution $\varepsilon_{\textup{F}}^{t}$ and the $\alpha$-JS divergence of the distribution of $\Phi$ between the treatment and control groups, up to some constants:
\begin{small}
\begin{align}
\varepsilon_{\PEHE}(f,\Phi)\leq 2\cdot (\varepsilon_{\textup{F}}^{0}&(f,\Phi)+\varepsilon_{\textup{F}}^{1}(f,\Phi)+\\\notag
&C_{\Phi,\alpha}\cdot \JSD_{\alpha}(p^{1}_{\Phi},p^{0}_{\Phi})-2\sigma_{Y}^{2}),
\end{align}
\end{small}
\hspace{-0.4em}where $C_{\Phi,\alpha}>0$ is a constant depending on the representation $\Phi$ and $\alpha$, and $\sigma_{Y}^{2}=\max_{t=0,1} E_{X}[\{(Y_{i}(t)-E(Y_{i}(t)|X))^{2}|X\}]$ is the expected conditional variance of $Y_{i}(t)$.
\end{theorem}
\hspace{-0.1em}The third theorem shows that the objective function in \eqref{eq:OF_EB} is an upper bound to the loss for the ITE estimation, which cannot be estimated based on the observed data. This theorem justifies the use of entropy as the distance metric in bounding the ITE prediction error.

\section{EXPERIMENTS}
We evaluate the proposed DRRL on the fully synthetic or semi-synthetic 
benchmark datasets. The experiment validates the use of DRRL and reveals 
several crucial properties of the representation learning for counterfactual 
prediction, such as the trade-off between balance and prediction power. The 
experimental details can be found in SM and the code is available from 
\url{https://github.com/zengshx777/Double-Rouble-Representation-Learning/}.
\vspace{-0.5em}
\subsection{Experimental Setups}
\vspace{-0.5em}
\label{sec:exp}

\textbf{Hyperparameter tuning, architecture} \quad As we only know one of the potential outcomes for each unit, we cannot perform hyperparameter selection on the validation data to minimize the loss. We tackle this problem in the same manner as \cite{shalit2017estimating}. Specifically, we use the one-nearest-neighbor matching method \citep{abadie2006matching} to estimate the ITE for each unit, which serves as the ground truth to approximate the prediction loss.
We use fully-connected multi-layer perceptrons (MLP) with ReLU activations as the flexible learner. The hyperparameters to be selected in the algorithm include the architecture of the network (number of representation layer, number of nodes in layer), the importance of imbalance measure $\kappa$, batch size in each learning step. We provide detailed hyperparameter selection steps in SM.

\textbf{Datasets} \quad To explore the performance of the proposed method extensively, we select the following three datasets: (i) \textbf{IHDP} \citep{hill2011bart,shalit2017estimating}: a semi-synthetic benchmark dataset with known ground-truth. The train/validation/test splits is 63/27/10 for 1000 realizations;(ii) \textbf{JOBS} \citep{lalonde1986evaluating}: a real-world benchmark dataset with a randomized study and an observational study. The outcome for the Jobs dataset is binary, so we add a sigmoid function after the final layer to produce a probability prediction and use the cross-entropy loss in \eqref{eq:OF_EB};
(iii) high-dimensional dataset, \textbf{HDD}: a fully-synthetic dataset with high-dimensional covariates and varying levels of confoundings. We defer its generating mechanism to Sec \ref{sec:high-dims}.

\textbf{Evaluation metrics} \quad 
To measure the performance of different counterfactual predictions algorithms, we consider the following evaluation metrics for both average causal estimands (including ATE and ATT) and ITE: (i) the absolute bias for ATE or ATT predictions $\varepsilon_{\ATE}=|\hat{\tau}_{\ATE}-\tau_{\ATE}|,\varepsilon_{\ATT}=|\hat{\tau}_{\ATT}-\tau_{\ATT}|$; (ii) the prediction loss for ITE, $\varepsilon_{\PEHE}$; (iii) \emph{policy risk}, quantifies the effectiveness of a policy depending on the outcome function $f_{t}(x)$, $R_{\textup{POL}}\triangleq 1-E(Y_{i}(1)|\pi_{f}(X_{i})=1)p(\pi_{f}=1)-E(Y_{i}(1)|\pi_{f}(X_{i})=0)p(\pi_{f}=0)$. It measures the risk of the policy $\pi_{f}$, which assigns treatment $\pi_{f}=1$ if $f_{1}(x)-f_{0}(x)>\delta$ and remains as control otherwise.

\textbf{Baselines} \quad  We compare DRRL with the following state-of-the-art methods: ordinary least squares (OLS) with interactions, k-nearest neighbor (k-NN), Bayesian Additive Regression Trees (BART) \citep{hill2011bart}, Causal Random Forests (Causal RF) \citep{wager2018causalforest}, Counterfactual Regression with Wasserstein  distance (CFR-WASS) or Maximum Mean Discrepancy (CFR-MMD) and their variant without balance regularization, the Treatment-Agnostic Representation Network(TARNet) \citep{shalit2017estimating}. We also evaluate the models that separate the weighting and representation learning procedure. Specifically, we replace the distance metrics in \eqref{eq:OF_EB} with other metrics like MMD or WASS, and perform entropy balancing on the learned representations (EB-MMD or EB-WASS).
\vspace{-1em}
\subsection{Learned Balanced Representations}
\vspace{-0.5em}

\begin{figure}[ht]
\vskip -0.1in
\begin{center}
\centerline{\includegraphics[width=1.0\columnwidth]{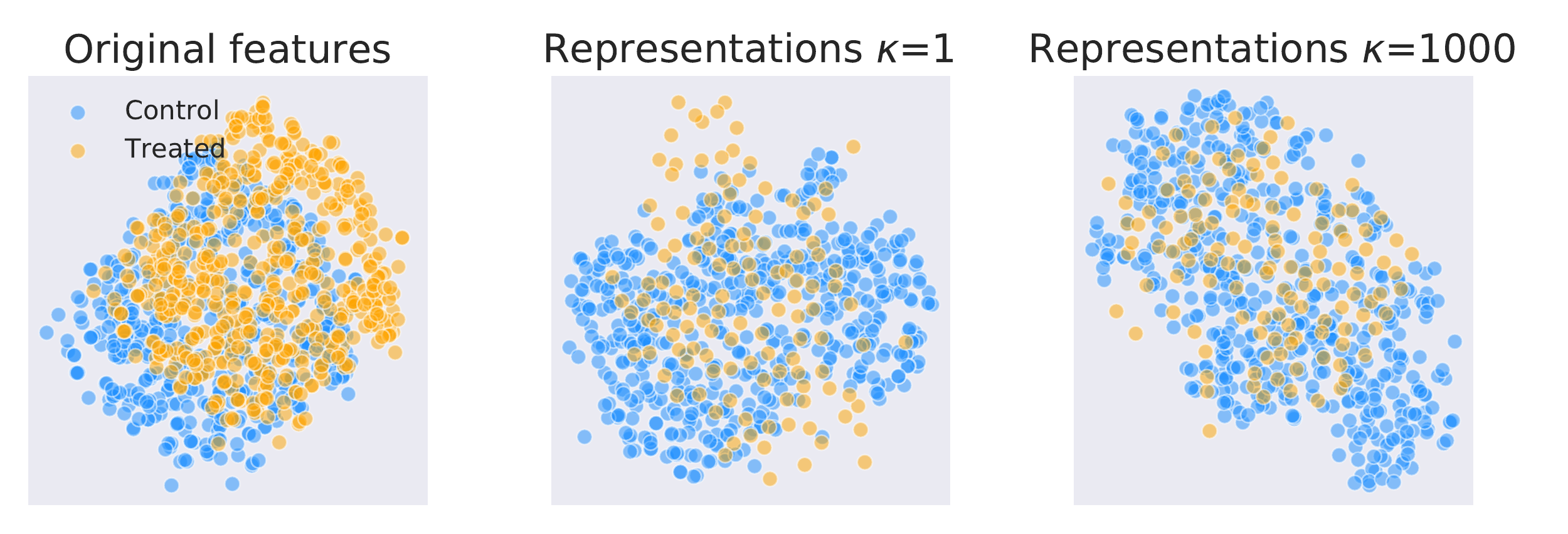}}
\caption{t-SNE visualization of original features, representations by DRRL when setting $\kappa=1,1000$.}
\label{fig:sne}
\end{center}
\vskip -0.2in
\end{figure}

We first examine how DRRL extracts balanced representations to support counterfactual predictions. In Figure \ref{fig:sne}, we select one imbalanced case from IHDP dataset and perform t-SNE (t-Distributed Stochastic Neighbor Embedding) \citep{maaten2008visualizing} to visualize the distribution of the original feature space and the representations learned from DRRL algorithm when $\kappa=1,1000$. 
While the original covariates are imbalanced, the learned representations or the transformed features have more similarity in distributions across two arms. Especially, a larger $\kappa$ value leads the algorithm to emphasize on the balance of representations and gives rise to a nearly identical representations across two groups. However, an overly large $\kappa$ may deteriorate the performance, because the balance is improved at the cost of predictive power. 

\begin{figure}[ht]
\vskip -0.2in
\begin{center}
\centerline{\includegraphics[width=1.0\columnwidth]{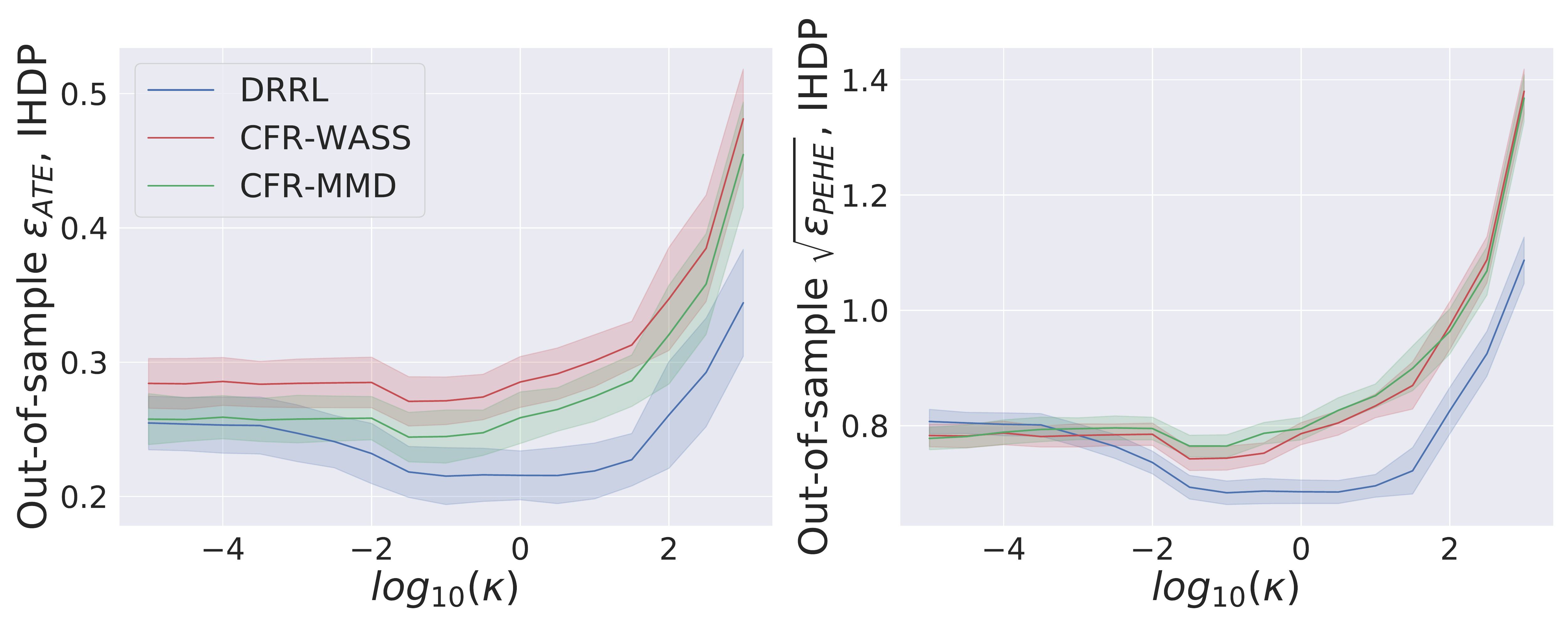}}
\caption{The sensitivity against the relative importance of balance $\kappa$ of $\varepsilon_{\ATE}$ (left) and $\varepsilon_{\PEHE}$ (right). Lower is better.}
\label{fig:sentivity}
\end{center}
\vspace{-2em}
\end{figure}

To see how the importance of balance constraint affects the prediction performance, we plot the $\varepsilon_{\ATE}$ and $\varepsilon_{\PEHE}$ in IHDP dataset against the hyperparameter $\kappa$ (on log scale) in Figure \ref{fig:sentivity}, for CFR-WASS, CFR-MMD and DRRL, which involve tuning $\kappa$ in the algorithms. We obtain the lowest $\varepsilon_{\ATE}$ or $\varepsilon_{\PEHE}$ at the moderate level of balance for the representations. This pattern makes sense as the perfect balance might compromise the prediction power of representations, while the poor balance cannot adjust for the confoundings sufficiently. Also, the DRRL is less sensitive to the choice $\kappa$ compared with CFR-WASS and CFR-MMD, with as the prediction loss has a smaller variation for different $\kappa$.

\begin{table*}[ht]
\centering
\caption{\label{tab:benchmarkresults}Results on IHDP datasets with 1000 replications, JOBS data and HDD dataset with 100 replications, average performance and its standard deviations. The models parametrized by neural network are in bold fonts}
\resizebox{2\columnwidth}{!}{
\begin{tabular}{lcc|cc|cccccc}
\hline
&\multicolumn{2}{c|}{\textbf{IHDP}}	&\multicolumn{2}{c}{\textbf{JOBS}}&\multicolumn{2}{c}{\textbf{HDD-A}}& \multicolumn{2}{c}{\textbf{HDD-B}}&\multicolumn{2}{c}{\textbf{HDD-C}}\\
&$\varepsilon_{\ATE}$&$\sqrt{\varepsilon_{\PEHE}}$&$\varepsilon_{\ATT}$&$R_{\textup{POL}}$&$\varepsilon_{\ATE}$&$\sqrt{\varepsilon_{\PEHE}}$&$\varepsilon_{\ATE}$&$\sqrt{\varepsilon_{\PEHE}}$&$\varepsilon_{\ATE}$&$\sqrt{\varepsilon_{\PEHE}}$\\
\hline\
OLS&$0.96\pm.06$&$6.6\pm .32$ &$0.08\pm .04$&$0.27\pm .03$&$-$&$-$&$-$&$-$&$-$&$-$\\
k-NN &$0.48\pm .04$&$3.9\pm .66$ &$0.11\pm .04$&$0.27\pm .03$ & $1.53\pm .14$& $7.71\pm .36$&$1.56\pm .18$&$6.94\pm .39$ &$1.78 \pm .23$& $6.95 \pm .40$\\
BART&$0.36\pm .04$&$3.2\pm .39$&$0.08\pm .03$&$0.28\pm .03$ & $0.97\pm .03$& $5.63\pm  .28$&$0.98\pm .06$&$4.31\pm .28$ &$0.94 \pm .08$& $3.94\pm .31$\\
Causal RF &$0.36\pm .03$&$4.0\pm .44$ &$0.09\pm .03$&$0.24\pm .03$ & $0.85\pm .05$& $5.52\pm .16$&$0.93\pm .05$&$4.14\pm .20$ &$0.87 \pm .06$& $3.17 \pm .27$\\
\textbf{TARNet}& $0.29\pm .02$& $0.94\pm .03$&$0.10\pm .03$& $0.28\pm .03$ &$1.05\pm .06$&$4.78\pm .16$ &$1.30\pm .08$ &$3.02\pm .17$ &$1.28\pm .09$ & $3.28 \pm .23$\\
\textbf{CFR-MMD}&$0.25\pm.02$&$0.76\pm.02$&$0.08\pm .03$&$0.26\pm .03$&$1.12\pm 
.05$&$4.45\pm .15$&$1.24\pm .05$&$2.71\pm.16$&$1.21 \pm .08$& $3.03 \pm .20$\\
\textbf{CFR-WASS}&$0.27\pm .02$& $0.74\pm.02$&$0.08\pm .03$&$0.27\pm 
.03$&$1.11\pm .06$&$4.48\pm .14$ &$1.15\pm .07$& $2.92 \pm .16$ &$1.22 \pm 
.08$& $2.91 \pm .19$\\
\textbf{EB-MMD} &$0.30\pm.02$ & $0.76\pm.03$ &$0.04\pm .01$ &$0.26\pm 
.03$&$1.07\pm .05$&$4.45\pm .15$&$0.98\pm .05$&$2.71\pm.16$&$1.00 \pm .08$& 
$3.03 \pm .20$\\
\textbf{EB-WASS }&$0.29\pm.02$ &$0.78\pm.03$  & $0.04\pm .01$ &$0.27\pm .03$ 
&$1.05\pm .06$&$4.48\pm .14$ &$1.03\pm .07$& $2.92 \pm .16$ &$1.02 \pm .08$& 
$2.91 \pm .19$\\
\textbf{DRRL}&$\mathbf{0.21\pm .03}$&$\mathbf{0.68\pm.02}$&$\mathbf{0.03\pm .02}$&$\mathbf{0.25\pm .02}$&$\mathbf{1.01\pm .04}$&$4.53\pm .15$&$ \mathbf{0.96\pm.04}$&$\mathbf{2.70\pm .16}$&$\mathbf{0.88 \pm .06}$& $\mathbf{2.57 \pm .17}$\\
\hline
\end{tabular}
}
\vskip -0.1in
\end{table*}

\vspace{-0.5em}
\subsection{Performance on Semi-synthetic or Real-world Dataset}
\vspace{-0.5em}
\textbf{ATE estimation} \quad We can see a significant gain in ATE estimation of DRRL over most state-of-the-art algorithms in the IHDP data, as in Table \ref{tab:benchmarkresults}; this is expected, as DRRL is designed to improve the inference of average estimands. The advantage remains even if we shift to binary outcome and the ATT estimand in the JOBS data, as in Table \ref{tab:benchmarkresults}. Moreover, compared with EB-MMD or EB-WASS which separates out the weights learning and representation learning, the proposed DRRL also achieve a lower bias in estimating ATE. This demonstrates the benefits of learning the weights and representation jointly instead of separating them out.

\begin{figure}[ht]
\centering
\vspace{-0.5em}
\includegraphics[width=0.9\columnwidth]{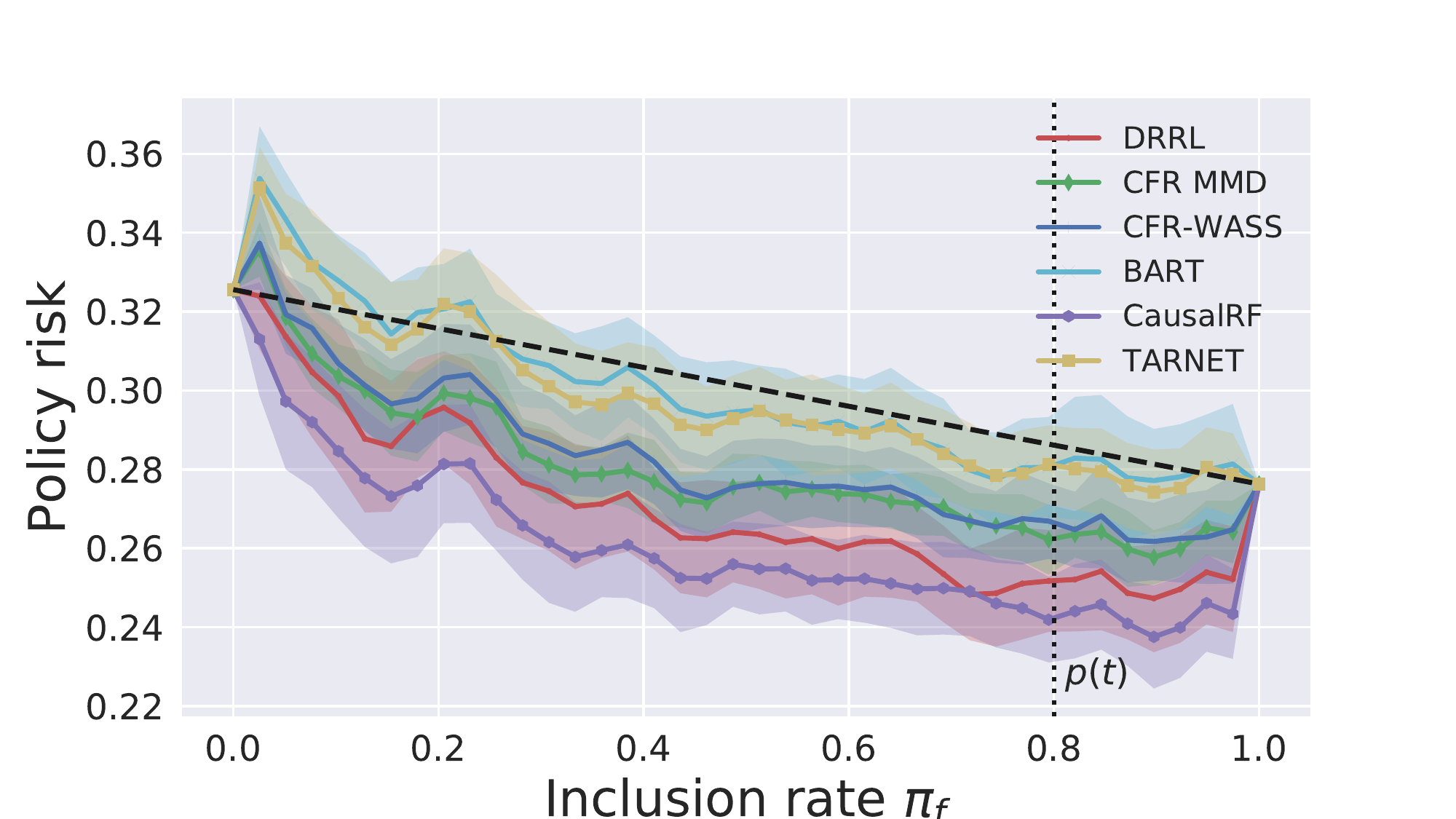}
\caption{The policy risk curve for different methods, using the random policy as a benchmark (dashed line). Lower value is better.}
\label{fig:policy_risk}
\vspace{-1em}
\end{figure}

\textbf{ITE estimation} \quad The DRRL has a better performance compared with the state-of-the-art methods like CFR-MMD on the IHDP dataset for ITE prediction. For the binary outcome in the JOBS data, the DRRL gives a better $R_{\textup{ROL}}$ over most methods except for the Causal RF when setting threshold $\delta=0$. In Figure \ref{fig:policy_risk}, we plot the policy risk as a function of the inclusion rate $p(\pi_{f}=1)$, through varying the threshold value $\delta$. The straight dashed line is the random policy assigning treatment with probability $\pi_{f}$, serving as a baseline for the performance. The vertical line shows the $\pi_{f}$ when $\delta=0$. The DRRL are persistently gives a lower $R_{\textup{ROL}}$  as we vary the inclusion rate of the policy


\subsection{High-dimensional Performance and Double Robustness}
\label{sec:high-dims}
We generate HDD datasets from the following model:
\begin{small}
\begin{gather*}
X_{i}\sim \mathcal{N} (0,\sigma^{2}[(1-\rho)I_{p}+\rho 1_{p}1_{p}^{T}])\\
||\beta_{0}||_{0}=||\beta_{\tau}||_{0}=||\gamma||_{0}=p^{\ast},\textup{supp}(\beta_{0})=\textup{supp}(\beta_{\tau})\\
P(T_{i}=1)=\textup{sigmoid}(X_{i}\gamma)\\
Y_{i}(t)=X_{i}\beta_{0}+TX_{i}\beta_{\tau}+\varepsilon_{i},\varepsilon_{i}\sim \mathcal{N}(0,\sigma_{e}^{2}),t=0,1,
\end{gather*}
\end{small}
\hspace{-0.4em}where $\beta_{0},\beta_{\tau},\gamma$ are the parameters for outcome and treatment assignment model. We consider sparse cases where the number of nonzero elements in $\beta_{0},\beta_{\tau},\gamma$ is much smaller than the total feature size $p^{\ast}<<p$. The support for $\beta_{0},\beta_{\tau}$ is the same, for simplicity.


Three scenarios are considered, by varying the overlapping support of $\gamma$ and $\beta_{0},\beta_{\tau}$: (i) scenario A (high confounding), the set of the variables determining the outcome and treatment assignment are identical, $||\textup{supp}(\beta_{0})\cap \textup{supp}(\gamma)||_{0}=p^{\ast}$; (ii) scenario B (moderate confounding), these two sets have 50\% overlapping, $||\textup{supp}(\beta_{0})\cap \textup{supp}(\gamma)||_{0}=p^{\ast}/2$; scenario C (low confounding), these two sets do not overlap, $||\textup{supp}(\beta_{0})\cap \textup{supp}(\gamma)||_{0}=0$. We set $p=2000,p^{\ast}=20,\rho=0.3$ and generate the data of size $N=800$ each time, with 54/21/25 train/validation/test splits. We report the $\varepsilon_{\ATE}$ and $\varepsilon_{\PEHE}$ in Table \ref{tab:benchmarkresults}\footnote{We omit the OLS here as it is the true generating model.}. The DRRL obtains the lowest error in estimating ATE, except for the Causal RF and BART, and achieve comparable performance in predicting ITE in all three scenarios.

This experiment also demonstrates the superiority of double robustness. The advantage of DRRL increases as the overlap between the predictors in the outcome function and those in the propensity score diminishes (from Scenario A to C), especially for ATE estimation. This illustrates the benefit of double robustness: when the representation learning fails to capture the predictive features of the outcomes, entropy balancing offers a second chance of correction via sample reweighting.

\section{CONCLUSIONS}

\label{sec:conclusion}
We propose a novel framework to learn double-robust representations for counterfactual prediction with the high-dimensional data. By incorporating an entropy balancing stage in the representation learning process and quantifying the balance of the representations between groups with the entropy of the resulting weights, we provide robust and efficient causal estimates. Important directions for future research include exploring other balancing weights methods \citep{deville1992calibration,kallus2019generalized,zubizarreta2015stable}  and generalizing into the learning problem with panel data \citep{abadie2010synthetic}, sequential treatments \citep{robins2000marginal}, and survival outcomes \citep{cox2018analysis}.

\bibliography{DR-NIPS}
\bibliographystyle{bib_style}

\onecolumn
\aistatstitle{Supplementary Materials for ``Double Robust Representation Learning for Counterfactual Prediction''}

\section{THEOREM PROOFS}
In this section, we present the detailed proofs for the theoretical properties in Sec 3.3 in the main text,
\begin{lemma}
	\label{Lemma1}
	The optimal value of dual variable $\lambdav^{\EB}$ converges to maximum likelihood estimator $\lambdav^{\star}$ in (12) in probability.
\end{lemma}

\textbf{Proof:}\quad The following proposition is based on a given representation, therefore we treat $\Phi(\cdot)$ as a fixed function. With Karush-Kuhn-Tucker (KKT) conditions, we derive the first order optimiality condition of (11):
\begin{small}
	\begin{equation}
	\label{eq:kkt}
	\begin{aligned}
	\sum_{i=1}^{n}(1-T_{i})e^{\sum_{j=1}^{m} \lambdav_{m}\Phi_{j}(X_{i})}(\Phi_{j}(X_{j})-\bar{\Phi_{j}})=0\\
	\sum_{i=1}^{n}T_{i}e^{-\sum_{j=1}^{m} \lambdav_{m}\Phi_{j}(X_{i})}(\Phi_{j}(X_{j})-\bar{\Phi_{j}})=0,
	\end{aligned}
	\end{equation}
\end{small}

for $j=1,2,\cdots,m$. We rewrite the above conditions as estimating equations, let $a_{j}(X,T,r,\lambdav)=(1-T)e^{\sum_{j=1}^{m} \lambdav_{j}\Phi_{j}(X)}(\Phi_{j}(X)-r_{j}),b_{j}(X,T,m,\lambdav)=Te^{\sum_{j=1}^{m} \lambdav_{j}\Phi_{j}(X)}(\Phi_{j}-r_{j})$. Then \eqref{eq:kkt} is the same as:
\begin{small}
	\begin{equation}
	\label{estimation_equations}
	\begin{aligned}
	\sum_{i}^{n}a_{j}(X_{i},T_{i},r,\lambdav)=0,j=1,2,\cdots,m\\
	\sum_{i}^{n}b_{j}(X_{i},T_{i},r,\lambdav)=0,j=1,2,\cdots,m.
	\end{aligned}
	\end{equation}
\end{small}
We can verify that $r_{j}=E(\Phi_{j}(X))$ and $\lambdav^{\ast}$ is the solution to the population version of \eqref{estimation_equations}. First, set $r_{j}=E(\Phi_{j}(X))$ and taking the conditional expectation of $a_{j},b_{j}$ given $X$ is:
\begin{small}
	\begin{equation}
	\begin{aligned}
	E(a_{j}(X,T,\lambdav,r)|X)=&\\
	(1-e(X))&e^{\sum_{j=1}^{m}\lambdav_{j}\Phi_{j}(X)}(\Phi_{j}(X)-E(\Phi_{j}(X)),\\
	E(b_{j}(X,T,\lambdav,r)|X)=&\\
	e(X)&e^{\sum_{j=1}^{m}-\lambdav_{j}\Phi_{j}(X)}(\Phi_{j}(X)-E(\Phi_{j}(X)).
	\end{aligned}
	\end{equation}
\end{small}
Suppose we are fitting the propensity score model with the log likelihood in (13), let $\lambdav^{\ast}=(\lambdav_{1}^{\ast},\cdots \lambdav_{m}^{\ast})$ be the MLE solution in (13) and plug into the $e(X)$, we have:
\begin{small}
	\begin{equation}
	\begin{aligned}
	E(a_{j}(X,T,\lambdav,r)|X)=\frac{e^{\sum_{j=1}^{m}\lambdav_{j}\Phi_{j}(x)}}{e^{\sum_{j=1}^{m}\lambdav^{\ast}_{j} \Phi_{j}(X)}}(\Phi_{j}(X)-E(\Phi_{j}(X)),\\
	E(b_{j}(X,T,\lambdav,r)|X)=\frac{e^{\sum_{j=1}^{m}-\lambdav_{j}\Phi_{j}(x)}}{e^{\sum_{j=1}^{m}-\lambdav^{\ast}_{j} \Phi_{j}(x)}}(\Phi_{j}(X)-E(\Phi_{j}(X)).
	\end{aligned}
	\end{equation}
\end{small}
The only way to make the follow quantify to be zero is to set $\lambdav_{j}=\lambdav_{j}^{\ast}$. So far we have verified $\lambdav^{\ast}$ is the solution to the population version of  $\eqref{estimation_equations}$, whose sample version is the KKT condition. Therefore, according to the M-estimation theory, we show that $\lambdav^{\EB}$ to $\lambdav^{\ast}$, which is the MLE solution for (13).

\begin{theorem}
The Shannon entropy of the EB weights defined in (4) converges in probability to the following $\alpha$-Jensen-Shannon divergence between the marginal representation distributions of the respective treatment groups:
\begin{small}
\begin{equation}\label{eq:EB_KL}
\begin{aligned}
\lim_{n \rightarrow \infty} \BH_n^{\EB}(\Phi)\triangleq \sum_{i}w_{i}^{\EB}(\Phi)\log (w_{i}^{\EB}(\Phi))&\stackrel{p}{\longrightarrow} \\ c'\{\KL(p_{\Phi}^{1}(x)||p_{\Phi}(x))+&\KL(p_{\Phi}^{0}(x)||p_{\Phi}(x))\}+c''=c' \JSD_{\alpha}(p_{\Phi}^{1},p_{\Phi}^{0})+c'' 
\end{aligned}
\end{equation}
\end{small}
where $c'>0, c''$ are non-zero constants,  $p_{\Phi}^{t}(x)=P(\Phi(\mathbf{X}_{i}=x)|T_{i}=t)$ is representation distribution in group $t$ ($t=0,1$), $p_{\Phi}(x)$ is the marginal density of the representations, $\alpha$ is the proportion of treated units $P(T_{i}=1)$ and $\KL(\cdot||\cdot)$ is the Kullback–Leibler (KL) divergence.
\end{theorem}
\textbf{Proof:}\quad  According to Lemma \ref{Lemma1}, we have $\lambdav^{\EB}\rightarrow \lambdav^{\ast}$. Therefore, with $w_{i}^{\EB}=\frac{\exp(-(2T_{i}-1)\sum_{j=1}^{m} \lambdav_{j}^{\EB}\Phi_{j}(X_{i})) }{\sum_{T_{i}=0}\exp(-(2T_{i}-1)\sum_{j=1}^{m} \lambda_{j}^{\EB}\Phi_{j}(X_{i}))}$, $w_{i}\propto \frac{1}{e(x_{i})}$ for $T_{i}=1$ and $w_{i}\propto \frac{1}{1-e(x_{i})}$ (up to a normalized constant) for $T_{i}=0$. Also, we have,
\begin{small}
	\begin{equation}
	\begin{aligned}
	\frac{1}{e(x_{i})}=\frac{1}{p(T_{i}=1|\Phi(X_{i}))}=\frac{p(\Phi(X_{i}))}{p(\Phi(X_{i})|T_{i}=1)p(T_{i}=1)},\\
	\frac{1}{1-e(x_{i})}=\frac{1}{p(T_{i}=0|\Phi(X_{i}))}=\frac{p(\Phi(X_{i}))}{p(\Phi(X_{i})|T_{i}=0)p(T_{i}=0)}.
	\end{aligned}
	\end{equation}
\end{small}
Also, we can derive,
\begin{small}
	\begin{equation}
	\begin{aligned}
	N_{1}w_{i}^{\EB}\rightarrow \frac{1/e(x_{i})}{\sum_{T_{i}=1}\frac{1}{e(x_{i})}/N_{1}}=\frac{1}{e(x_{i})}/E(1/e(x_{i})|T_{i}=1).
	\end{aligned}
	\end{equation}
\end{small}
Notice that,
\begin{small}
	\begin{equation}
	\begin{aligned}
	E(1/e(x_{i})|T_{i}=1)&=\int_{\mathcal{X}} \frac{1}{e(x)}p(x|T=1)dx\\
	&=\int_{\mathcal{X}}\frac{p(\Phi(X_{i}))}{p(T=1)}dx=p(T_{i}=1).\\
	\end{aligned}
	\end{equation}
\end{small}
Therefore, we have,
\begin{small}
	\begin{equation}
	\begin{aligned}
	N_{1}w_{i}^{\EB}\rightarrow \frac{p(\Phi(X_{i}))}{p(\Phi(X_{i})|T_{i}=1)},\\
	\end{aligned}
	\end{equation}
\end{small}
where $N_{1}$ is the number of treated. For the entropy of the EB weights,
\begin{small}
	\begin{equation}
	\begin{aligned}
	-\sum_{T_{i}=1}w_{i}^{\EB}&\log w_{i}^{\EB}=\frac{\sum_{T_{i}=1} N_{1}w_{i}^{\EB}\log N_{1}w_{i}^{\EB} }{N_{i}}+c_{1}''\\
	&=E_{x_{i}|T_{i}=1}(N_{1}w_{i}^{\EB}\log N_{1}w_{i}^{\EB}))+c_{1}''\\
	&=-\int_{\mathcal{X}}  \log \frac{p(\Phi(x))}{p(\Phi(x)|T_{i}=1)} p(\Phi(x)|T=1)dx+c_{1}''\\
	&=\int_{\mathcal{X}}\log \frac{p(\Phi(x)|T=1)}{p(x)}p(\Phi(x)|T=1)dx+c_{1}''\\
	&=\KL(p(\Phi(x)|T=1)||p(\Phi(x)))+c_{1}''.
	\end{aligned}
	\end{equation}
\end{small}
Similarly, we have,
\begin{small}
	\begin{eqnarray*}
		-\sum_{T_{i}=0}w_{i}^{\EB}\log w_{i}^{\EB}\rightarrow \KL(p(\Phi(x)|T=0)||p(\Phi(x)))+c_{0}''.
	\end{eqnarray*}
\end{small}
Therefore, we can conclude that 
\begin{small}
	\begin{equation}
	\begin{aligned}
	-\sum_{i}w_{i}&^{\EB}\log w_{i}^{\EB}\rightarrow \\ &c'[\KL(p_{\Phi}^{0}(x)||p_{\Phi}(x))+\KL(p_{\Phi}^{1}(x)||p_{\Phi}(x))]+c''.
	\end{aligned}
	\end{equation}
\end{small}
Specifically, with $p_{\Phi}(x)=\alpha p_{\Phi}^{1}(x)+(1-\alpha)p_{\Phi}^{0}(x)$, with $\alpha=p(T_{i}=1)$ we can conclude that 
\begin{small}
	\begin{eqnarray*}
		-\sum_{i}w_{i}^{\EB}\log w_{i}^{\EB}\rightarrow c'\JSD_{\alpha}(p_{\Phi}^{1}(x),p_{\Phi}^{0}(x))+c''.
	\end{eqnarray*}
\end{small}
Therefore, we show that the max entropy is a linear transformation of $\JSD_{\alpha}(p_{\Phi}^{1}(x)|p_{\Phi}^{0}(x))$. We can use its negative value $\sum_{i}w_{i}^{\EB}log w_{i}^{\EB}$ as a measure of balance.
\begin{theorem}[Double Robustness]
	\label{thm:DR} 
Under the strong ignorability, the entropy balancing estimator $\hat{\tau}_{\ATE}^{\EB}$ is consistent for $\tau_{\ATE}$  if
either the true outcome models $f_t(x,t),t\in\{0,1\}$ or the true propensity score model $\logit\{e(x)\}$ is linear in representation $\Phi(x)$.
\end{theorem}
\textbf{Proof:(a)Correctly specified propensity score model} \quad Suppose $\logit\{e(x)\}$ is linear in $\Phi(x)$, which means fitting a logistic regression between $T_{i}$ and $\Phi(x)$ is a correctly specified model for the propensity score. Therefore, according to Lemma \ref{Lemma1}, we have $\hat{w}_{i}^{\EB}\rightarrow \frac{1}{e(x_{i})}$ for $T_{i}=1$ and $\hat{w}_{i}^{\EB}\rightarrow\frac{1}{1-e(x_{i})}$ for $T_{i}=0$. The estimator in (8) can be expressed as,
\begin{small}
	\begin{equation}
	\begin{aligned}
	\label{ps_decom}
	\hat{\tau}_{\ATE}^{\EB}&=
	\sum_{T_{i}=1}\hat{w}_{i}^{\EB}Y_{i}-\sum_{T_{i}=0}\hat{w}_{i}^{\EB}Y_{i}+\\
	&\frac{1}{N}\sum_{i=1}^{N}(T_{i}\hat{w}_{i}^{\EB}N-1)\hat{f}_{1}(\hat{\Phi}(X_{i}))\\
	&-\frac{1}{N}\sum_{i=1}^{N}((1-T_{i})\hat{w}_{i}^{\EB}N-1)\hat{f}_{0}(\hat{\Phi}(X_{i})),
	\end{aligned}
	\end{equation}
\end{small}
where we $\sum_{T_{i}=1}\hat{w}_{i}^{\EB}Y_{i}-\sum_{T_{i}=0}\hat{w}_{i}^{\EB}Y_{i}$ converges to $\tau^{\ATE}$, which is the usual IPW estimator when the propensity score model is correctly specified. For the last two terms in \eqref{ps_decom},
\begin{small}
	\begin{equation}
	\begin{aligned}
	\sum_{i=1}^{N}(T_{i}\hat{w}_{i}^{\EB}N-1)&\hat{f}_{1}(\hat{\Phi}(X_{i}))\\
	&=
	N\sum_{T_{i}=1}\hat{w}_{i}^{\EB}\hat{\gamma}_{1}'\hat{\Phi}(X_{i})-N\sum_{i=1}^{N}\hat{\gamma}_{1}'\hat{\Phi}(X_{i})/N\\
	&=N\sum_{T_{i}=1}\sum_{j=1}^{m}\hat{\gamma}_{1j}(\hat{w}_{i}^{\EB}\hat{\Phi}_{j}(X_{i})-\bar{\Phi_{j}}(X_{i}))=0.
	\end{aligned}
	\end{equation}
\end{small}
The second equality follows from the balance constraint in (7). Similarly, we can show that $\frac{1}{N}\sum_{i=1}^{N}((1-T_{i})\hat{w}_{i}^{\EB}N-1)\hat{f}_{0}(\hat{\Phi}(X_{i}))=0$. Therefore, we have shown that $\hat{\tau}_{\ATE}^{\EB}$ converges to $\tau^{\ATE}$ when propensity score model is correctly specified.

\textbf{(b)Correctly specified outocme model}\quad Suppose the true outcome function is linear in representation $\Phi(x)$, thus $f(x,0)=\gamma_{0}'\Phi(x),f(x,1)=\gamma_{1}'\Phi(x)$, which means $\hat{f}_{1}(\hat{\Phi}(X_{i}))\rightarrow f_{1}(\Phi(X_{i})),\hat{f}_{0}(\hat{\Phi}(X_{i}))\rightarrow f_{0}(\Phi(X_{i}))$. Then we have,
\begin{small}
	\begin{align}
	\sum_{T_{i}=1}& \hat{w}_{i}^{\EB}\{Y_{i}-\hat{f}_{1}(\hat{\Phi}(X_{i}))\}\\\nonumber
	\rightarrow &E\{N_{1}\hat{w}_{i}^{\EB}(Y_{i}-f_{1}(\Phi(X_{i})))|T_{i}=1\}\\
	&=E\{N_{1}\hat{w}_{i}^{\EB}(E(Y_{i}|X_{i},T_{i}=1)-f_{1}(\Phi(X_{i})))\} \label{eq:LLE}\\
	&=E\{N_{1}\hat{w}_{i}^{\EB}(E(Y_{i}(1)|X_{i})-f_{1}(\Phi(X_{i}))) \}\label{eq:ignore}\\
	&=E\{N_{1}\hat{w}_{i}^{\EB}(f_{1}(\Phi(X_{i}))-f_{1}(\Phi(X_{i}))) \}
	=0.
	\end{align}
\end{small}
The first equality \eqref{eq:LLE} follows from the law of iterated expectation. The second equality \eqref{eq:ignore} follows from the ignorability assumption (1). Similarly, we can prove that,
\begin{small}
	\begin{eqnarray}
	\sum_{T_{i}=0} \hat{w}_{i}^{\EB}\{Y_{i}-\hat{f}_{0}(\hat{\Phi}(X_{i}))\}\rightarrow 0.
	\end{eqnarray}
\end{small}
Therefore, the first term in (8), $\sum_{i=1}^{N}\hat{w}_{i}^{\EB}(2T_{i}-1)\{Y_{i}-\hat{f}_{T_{i}}(\hat{\Phi}(X_{i}))\}=\sum_{T_{i}=1} \hat{w}_{i}^{\EB}\{Y_{i}-\hat{f}_{1}(\hat{\Phi}(X_{i}))\}+	\sum_{T_{i}=0} \hat{w}_{i}^{\EB}\{Y_{i}-\hat{f}_{0}(\hat{\Phi}(X_{i}))\}\rightarrow 0$. Also, the second term in (8) converges to the true $\tau^{\ATE}$,
\vspace{-5pt}
\begin{small}
	\begin{equation*}
	\begin{aligned}
	\frac{1}{N}\sum_{i=1}^{N}\{\hat{f}_{1}(\hat{\Phi}(X_{i}))&-\hat{f}_{0}(\hat{\Phi}(X_{i}))\}\rightarrow E(f_{1}(\Phi(X_{i}))-f_{0}(\Phi(X_{i}))\\
	&=E\{E(Y_{i}(1)|X_{i})-E(Y_{i}(0)|X_{i})\}\\
	&=E\{E((Y_{i}(1)-Y_{i}(0))|X_{i})\}\\
	&=E(Y_{i}(1)-Y_{i}(0))=\tau^{\ATE}
	\end{aligned}
	\end{equation*}
\end{small}
Based on the consistency under condition (a) and (b), we can conclude estimator in (8) is doubly robust for $\tau^{\ATE}$.

We list two lemmas required for the proof of Theorem \ref{thm:bound}. Lemma \ref{Lemma2} defines the counterfactual loss and show that the expected loss of estimating ITE can be bounded by the sum of factual loss and counterfactual loss.

\begin{lemma}
	\label{Lemma2}
	For given outcome function $f$ and representation $\Phi$, define the counterfactual loss for treatment arm $t$ as,
	\begin{small}
		\begin{eqnarray}
		\varepsilon_{\textup{CF}}^{t}(f,\Phi)=\int_{\mathcal{X}}l_{f,\Phi}(x,t)p_{1-t}(x)dx.
		\end{eqnarray}
	\end{small}
	Then, we can bound the expected loss in estimating $\varepsilon_{\PEHE}$ by the factual loss $\varepsilon_{\textup{F}}^{t}(f,\Phi)$ and counterfactual loss $\varepsilon_{\textup{CF}}^{t}(f,\Phi)$, 
	\begin{small}
		\begin{eqnarray}
		\varepsilon_{\PEHE}(f,\Phi)&\leq& 2(\varepsilon_{\textup{F}}(f,\Phi)+\varepsilon_{\textup{CF}}(f,\Phi)-2\sigma_{Y}^{2}),\\
		\varepsilon_{\textup{F}}(f,\Phi)&=&\alpha \varepsilon_{\textup{F}}^{1}(f,\Phi)+(1-\alpha)\varepsilon_{\textup{F}}^{0}(f,\Phi),\\
		\varepsilon_{\textup{CF}}(f,\Phi)&=&(1-\alpha) \varepsilon_{\textup{CF}}^{1}(f,\Phi)+\alpha\varepsilon_{\textup{CF}}^{0}(f,\Phi),
		\end{eqnarray}
	\end{small}
	where $\sigma_{Y}^{2}=\max_{t=0,1} E_{X}[\{(Y_{i}(t)-E(Y_{i}(t)|X))^{2}|X\}]$ is the expected conditional variance of  $Y_{i}(t)$ over the covariate space $\mathcal{X}$.
\end{lemma}
\textbf{Proof:}\quad 
This lemma is exactly the same as the proof for the first inequality of Theorem 1 in \cite{shalit2017estimating}. We refer readers to that part for conciseness. 

Lemma \ref{Lemma3} below outlines the connection between the total variation distance and $\alpha$-JS divergence.

\begin{lemma}
	\label{Lemma3}
	The total variational distance between distributions $p$ and $q$ can be bounded by the $\alpha$-JS divergence,
	\begin{small}
		\begin{equation}
		\begin{aligned}
		TV(p,q)=\int |p(x)-q(x)|dx&\leq \frac{2}{\alpha}\sqrt{(1-e^{-\JSD_{\alpha}(p,q)})}\\
		&\leq \frac{2}{\alpha}\sqrt{\JSD_{\alpha}(p,q)}.
		\end{aligned}
		\end{equation}
	\end{small}
\end{lemma}

\textbf{Proof:}\quad 
Define $r_{\alpha}(x)=(1-\alpha)p(x)+\alpha q(x)$, we evaluate $\KL(p(x)||r_{\alpha}(x))$,
\begin{small}
\begin{align}
	\KL&(p(x)||r_{\alpha}(x))=-\int p(x)\log \frac{r_{\alpha}(x)}{p(x)}\\
	&=-\int p(x)[\log \min(\frac{r_{\alpha}(x)}{p(x)},1)+\log \max(\frac{r_{\alpha}(x)}{p(x)},1)]dx\label{eq:minmaxeq}\\
	&\geq -\log \int p(x)\min(\frac{r_{\alpha}(x)}{p(x)},1)dx-\nonumber\\
	&\quad \log \int p(x)\max(\frac{r_{\alpha}(x)}{p(x)},1)dx\label{eq:jensen}\\
	&=-\log \int \min(r_{\alpha}(x),p(x))dx-\nonumber\\
	&\quad \log \int \max(r_{\alpha}(x),p(x))dx\\
	&=-\log \int (\frac{p(x)+r_{\alpha}(x)}{2}-\frac{|p(x)-r_{\alpha}(x)|}{2})dx\nonumber\\
	&\quad -\log \int (\frac{p(x)+r_{\alpha}(x)}{2}+\frac{|p(x)-r_{\alpha}(x)|}{2})dx\label{eq:abs_minmax}\\
	&=-\log (1-\frac{\alpha}{2}\int|p(x)-q(x)|dx)+\nonumber\\
	&\quad \log(1+\frac{\alpha}{2}\int|p(x)-q(x)|dx)\\
	&=-\log (1-\frac{\alpha^{2}}{4}TV^{2}(p,q)).
\end{align}
\end{small}
The second equality \eqref{eq:minmaxeq} follows from the fact that $x=\min(x,1)\max(x,1)$. The first inequality \eqref{eq:jensen} follows from Jensen inequality. The fourth equality \eqref{eq:abs_minmax} follows from the fact that $\min(a,b)=\frac{a+b}{2}-\frac{|a-b|}{2},\max(a,b)\frac{a+b}{2}+\frac{|a-b|}{2}$. 
which indicates,
\begin{small}
	\begin{align}
	\JSD_{\alpha}(p,q)&=\frac{1}{2}[\KL(p(x)||r_{\alpha}(x))+\KL(q(x)||r_{\alpha}(x))]\\
	&\geq -\log (1-\frac{\alpha^{2}}{4}TV^{2}(p,q)),\\
	TV(p,q)&\leq \frac{2}{\alpha}\sqrt{(1-e^{-\JSD_{\alpha}(p,q)})}\leq \frac{2}{\alpha}\sqrt{\JSD_{\alpha}(p,q)}.\label{TV_inequal}
	\end{align}
\end{small}
The second inequality in \eqref{TV_inequal} follows from the fact that $1-e^{-x}\leq x$.

With Lemma \ref{Lemma2} and \ref{Lemma3}, we proceed to prove Theorem \ref{thm:bound}. The strategy is to bound by the counterfactual loss by the factual loss and total variation distance. Next step, we replace total variation distance with $\alpha$-JS divergence. In the final, we bound the loss of estimating ITE by the counterfactual loss an factual loss with Lemma \ref{Lemma2}.
\begin{theorem}
	\label{thm:bound}
Suppose $\mathcal{X}$ is a compact space and $\Phi(\cdot)$ is a continuous and invertible function. For a given $f,\Phi$, the expected loss for estimating the ITE, $\varepsilon_{\textup{PEHE}}$, is bounded by the sum of the prediction loss on the factual distribution $\varepsilon_{\textup{F}}^{t}$ and the $\alpha$-JS divergence of the distribution of $\Phi$ between the treatment and control groups, up to some constants:
\begin{small}
\begin{equation}
\varepsilon_{\PEHE}(f,\Phi)\leq 2\cdot (\varepsilon_{\textup{F}}^{0}(f,\Phi)+\varepsilon_{\textup{F}}^{1}(f,\Phi)+C_{\Phi,\alpha}\cdot \JSD_{\alpha}(p^{1}_{\Phi},p^{0}_{\Phi})-2\sigma_{Y}^{2}),
\end{equation}
\end{small}
where $C_{\Phi,\alpha}>0$ is a constant depending on the representation $\Phi$ and $\alpha$, and $\sigma_{Y}^{2}=\max_{t=0,1} E_{X}[\{(Y_{i}(t)-E(Y_{i}(t)|X))^{2}|X\}]$ is the expected conditional variance of $Y_{i}(t)$.
\end{theorem}

\textbf{Proof:}\quad Let $\Psi(\cdot):\mathcal{R}^{m}\rightarrow \mathcal{X}$ denote the inverse mapping of $\Phi(X)$. First, we bound the counterfactual loss $\varepsilon_{\textup{CF}}(f,\Phi)$ with the factual loss $\varepsilon_{\textup{F}}(f,\Phi)$ and $\alpha$-JS divergence,
\begin{small}
\begin{align}
	|\varepsilon_{\textup{CF}}^{0}(f,\Phi)&-\varepsilon_{\textup{F}}^{0}(f,\Phi)|\\
	&=|\int_{\mathcal{X}}l_{f,\Phi}(x,0)p_{1}(x)dx-\int_{\mathcal{X}}l_{f,\Phi}(x,0)p_{0}(x)dx|\\
	&\leq \int_{\mathcal{X}}l_{f,\Phi}(x,0)|p_{1}(x)-p_{0}(x)|dx\\
	&=\int_{\mathcal{R}^{m}}l_{f,\Phi}(\Psi(s),0)|p_{\Phi}^{1}(s)-p_{\Phi}^{0}(s)|ds\label{eq:changevf}\\
	&\leq B_{\Phi}\int_{\mathcal{R}^{m}}|p_{\Phi}^{1}(s)-p_{\Phi}^{0}(s)|ds
	=B_{\Phi}TV(p_{\Phi}^{1},p_{\Phi}^{0})\label{eq:bound}\\
	&\leq \frac{2B_{\Phi}}{\alpha}\sqrt{\JSD_{\alpha}(p_{\Phi}^{1},p_{\Phi}^{0})}.\label{eq:apply_lemma3}
\end{align}
\end{small}
The equality \eqref{eq:changevf} follows from the change of variable formula, the second inequality \eqref{eq:bound} from the fact that $l_{f,\Phi}(\Psi(s),0)$ is a continuous function on a compact space. The third inequality \eqref{eq:apply_lemma3} follow from Lemma \ref{Lemma3}. With similar argument, we can derive that,
\begin{small}
	\begin{eqnarray}
	|\varepsilon_{\textup{CF}}^{1}(f,\Phi)-\varepsilon_{\textup{F}}^{1}(f,\Phi)|\leq 	\frac{2B_{\Phi}'}{\alpha} \sqrt{\JSD_{\alpha}(p_{\Phi}^{1},p_{\Phi}^{0})}.
	\end{eqnarray}
\end{small}

Therefore, we have,
\begin{small}
	\begin{equation}
	\begin{aligned}
	|\varepsilon_{\textup{CF}}&(f,\Phi)-\alpha \varepsilon_{\textup{F}}^{0}(f,\Phi)+(1-\alpha)\varepsilon_{\textup{F}}^{1}(f,\Phi)|\\
	&=\alpha|\varepsilon_{\textup{CF}}^{0}(f,\Phi)-\varepsilon_{\textup{F}}^{0}(f,\Phi)|+(1-\alpha)|\varepsilon_{\textup{CF}}^{1}(f,\Phi)-\varepsilon_{\textup{F}}^{1}(f,\Phi)|\\
	&\leq 2\frac{(1-\alpha)B_{\Phi}+\alpha B_{\Phi}'}{\alpha} \sqrt{\JSD_{\alpha}(p_{\Phi}^{1},p_{\Phi}^{0})}\\
	&\leq C_{\Phi,\alpha}\sqrt{\JSD_{\alpha}(p_{\Phi}^{1},p_{\Phi}^{0})}\\.
	\end{aligned}
	\end{equation}
\end{small}
With Lemma \ref{Lemma2}, we have 
\begin{small}
	\begin{equation}
	\begin{aligned}
	\varepsilon_{\PEHE}&(f,\Phi)\leq 2(\varepsilon_{\textup{F}}(f,\Phi)+\varepsilon_{\textup{CF}}(f,\Phi)-2\sigma_{Y}^{2})\\
	&\leq 2(\alpha \varepsilon_{\textup{F}}^{1}(f,\Phi)+(1-\alpha)\varepsilon_{\textup{F}}^{0}(f,\Phi)+\varepsilon_{\textup{CF}}(f,\Phi)-2\sigma_{Y}^{2})\\
	&\leq  2 (\alpha \varepsilon_{\textup{F}}^{1}(f,\Phi)+(1-\alpha)\varepsilon_{\textup{F}}^{0}(f,\Phi)+\alpha \varepsilon_{\textup{F}}^{0}(f,\Phi)\\
	&+(1-\alpha)\varepsilon_{\textup{F}}^{1}(f,\Phi)+C_{\Phi,\alpha}\sqrt{\JSD_{\alpha}(p_{\Phi}^{1},p_{\Phi}^{0})}-2\sigma_{Y}^{2})\\
	&=2(\varepsilon_{\textup{F}}^{0}(f,\Phi)+\varepsilon_{\textup{F}}^{1}(f,\Phi)+C_{\Phi,\alpha}\sqrt{\JSD_{\alpha}(p_{\Phi}^{1},p_{\Phi}^{0})}-2\sigma_{Y}^{2}),
	\end{aligned}
	\end{equation}
\end{small}
which proves the inequality in Theorem \ref{thm:bound} (typo correction: missing squared root in the main text on $\JSD_{\alpha}(p_{\Phi}^{1},p_{\Phi}^{0})$).
\section{GENERALIZATION TO OTHER ESTIMANDS}
In this section, we use $\tau^{\ATT}$ an example of how to generalize to other estimands. If we are interested in estimating  $\tau^{\ATT}$, we can solve the following optimization problem,
\begin{small}
\begin{equation}
\label{eq:entropy_balancing_representation_att}
\begin{aligned}
\max_{w} \quad &-\sum_{T_{i}=0}^{N}w_{i}\log w_{i},\\
\textrm{s.t} \quad &\sum_{T_{i}=0}w_{i} \Phi_{ji}=\sum_{T_{i}=1}\Phi_{ji}/N_{1}=\bar{\Phi}_{j}(1),j=1,2\cdots,m, \\
&\sum_{T_{i}=0}w_{i}=1, w_{i}>0.\\
\end{aligned}
\end{equation}
\end{small}
And our estimator for $\tau^{\ATT}$ is 
\begin{small}
\begin{eqnarray}
\hat{\tau}_{\ATT}^{\EB}=\sum_{T_{i}=1}Y_{i}/N_{1}-\sum_{T_{i}=0}\hat{w}_{i}^{\EB}Y_{i}.
\end{eqnarray}
\end{small}
We prove its double robustness in Theorem \ref{thm:DR_ATT}. 

\begin{theorem}[Double Robustness for ATT]
\label{thm:DR_ATT} 
Under Assumptions 1 and 2, the entropy balancing estimator $\hat{\tau}_{\ATT}^{\EB}$ with the weights $w_{i}^{\EB}(\Phi)$ solved from Problem (6) and \eqref{eq:entropy_balancing_representation_att} is doubly robust in the sense that: 
If either the true outcome model $f(x,0)1$ or the true propensity score model $\logit\{e(x)\}$ is linear in the representations $\Phi(x)$, then $\hat{\tau}_{\ATT}^{\EB}$ is consistent for $\tau_{\ATT}$.
\end{theorem}
\textbf{Proof:}\quad The dual problem for the optimization problem is 
\begin{small}
\begin{equation}
\label{eq:dual_entropy_att}
\begin{aligned}
\min_{\lambdav} \quad & \log(\sum_{T_{i}=0}exp(\sum_{j=1}^{m}\lambdav_{j}\Phi_{j}(X_{i})))-\sum_{j=1}^{m}\lambdav_{j}\bar{\Phi}_{j}(1)\\
\end{aligned}
\end{equation}
\end{small}
where $\lambdav_{j}$ is the Lagrangian multiplier. With KKT condition, the optimal weights are
\begin{small}
\begin{eqnarray}
\label{eq:softmax}
w_{i}^{\EB}=\frac{exp(\sum_{j=1}^{m} \lambdav_{j}^{\EB}\Phi_{j}(X_{i})) }{\sum_{T_{i}=0}exp(\sum_{j=1}^{m} \lambdav_{j}^{\EB}\Phi_{j}(X_{i}))}
\end{eqnarray}
\end{small}
where $\lambdav^{\EB}$ is the solution to the dual problem \eqref{eq:dual_entropy_att}. 
\textbf{(a)Correctly specified propensity score model} \quad If the logit of true propensity score value, $\log(\frac{e(X_{i})}{1-e(X_{i})})$ is linear in $\Phi_{j}(X_{i})$, then we can show that $\lambdav^{\EB}$ converges to the solution $\lambdav^{\ast}$ to the following optimization problem by Lemma \ref{Lemma1}.
\begin{small}
\begin{equation}
\label{eq:logistic_regression}
\begin{aligned}
\min_{\lambdav} \quad & \sum_{T_{i}=0}\log(1+exp(-(2T_{i}-1)\sum_{j=1}^{m}\lambdav_{j}\Phi_{j}(X_{i})))\\
\end{aligned}
\end{equation}
\end{small}
which is maximizing the log likelihood when fitting a logistic regression between $T_{i}$ and $\Phi_{j}(X_{i})$. As long as we have $\lambdav^{\EB}$ converges to $\lambdav^{\ast}$, we can claim that $N_{0}w_{i}^{\EB}\rightarrow c\frac{e(X_{i})}{1-e(X_{i})}$\cite{zhao2017entropy}, $c$ is some normalized constant, which proves the consistency of the estimator.

\textbf{(b)Correctly specified outcome model}\quad If outcome model $f(x,0)$ is linear in $\Phi_{j}(X_{i})$, then we can expand $E(Y_{i}(0)|X_{i}=x)=f(x,0)=\sum_{j=1}^{m}\gamma_{0j}\Phi_{j}(x)$.
\begin{small}
\begin{equation}
\begin{aligned}
E(Y_{i}&(0)|T_{i}=1)\\
&=\int_{\mathcal{X}} E(Y_{i}(0)|X_{i}=x,T_{i}=1)p_{1}(x)dx,\\
&=\int_{\mathcal{X}}  E(Y_{i}(0)|X_{i}=x)p_{1}(x)dx,\\
&=\sum_{j=1}^{m}\gamma_{0j}\int \Phi_{j}(x)p_{1}(x)dx.
\end{aligned}
\end{equation}
\end{small}

We also have,
\begin{small}
\begin{equation}
\begin{aligned}
\sum_{T_{i}=0}w_{i}^{\EB}Y_{i}&=\sum_{T_{i}=0}w_{i}^{\EB'}Y_{i}/N_{0}\rightarrow E\{w_{i}^{\EB'}Y_{i}(0)|T_{i}=0\}\\
&=\int w_{i}^{\EB'}E(Y_{i}(0)|X_{i})p_{0}(x)dx\\
&=\sum_{j=1}^{m}\gamma_{0j}\int w_{i}^{\EB'}\Phi_{j}(x)p_{0}(x)dx.
\end{aligned}
\end{equation}
\end{small}
where $w_{i}^{\EB'}$ is the normalized $w_{i}^{\EB}$ with $w_{i}^{\EB'}=N_{0}w_{i}^{\EB}$. Notice that,
\begin{small}
\begin{equation}
\begin{aligned}
\sum_{T_{i}=0}w^{\EB'}\Phi_{j}(X_{i})/N_{0}&\rightarrow\int w_{i}^{\EB'}\Phi_{j}(x)p_{1}(x)dx,\\
\sum_{T_{i}=1}\Phi_{j}(X_{i})/N_{1}&\rightarrow \int \Phi_{j}(x)p_{1}(x)dx,
\end{aligned}
\end{equation}
\end{small}
By the constraints of \eqref{eq:entropy_balancing_representation_att}, we have:
\begin{small}
\begin{equation}
\begin{aligned}
\sum_{T_{i}=0}w^{'\EB}\Phi_{j}(X_{i})/N_{0}=\sum_{T_{i}=0}w^{\EB}\Phi_{j}(X_{i})=\sum_{T_{i}=1}\Phi_{j}(X_{i})/N_{1}.
\end{aligned}
\end{equation}
\end{small}
Therefore, we have
\begin{small}
\begin{equation}
\begin{aligned}
\int \Phi_{j}(x)p_{1}(x)dx=\int w_{i}^{\EB'}\Phi_{j}(x)p_{0}(x)dx,
\end{aligned}
\end{equation}
\end{small}
which implies
\begin{small}
 \begin{equation}
 \begin{aligned}
   \sum_{T_{i}=0}w_{i}^{\EB}Y_{i}\rightarrow E(Y_{i}(0)|T_{i}=1).
 \end{aligned}
 \end{equation}
\end{small}	
With $\sum_{T_{i}=1}Y_{i}/N_{1}\rightarrow E(Y_{i}(1)|T_{i}=1)$, we establish the consistency if outcome model is correctly specified.

Based on (a) and (b), we show $\hat{\tau}_{\ATT}^{\EB}$ is doubly robust. The proof is largely follows from \cite{zhao2017entropy}.

\section{EXPERIMENTAL DETAILS}
\subsection{HYPERPARAMETER SELECTION}
We random sample one combination from all possible choice hyperparameters and train the model on the experimental dataset each time.  We perform the hyperparameters selection regime described in section 6 and report only the best one within all possible choices in the random sampling. Table \ref{tab:hyperparameters} lists all possible choice for the parameter. For IHDP data and high-dimensional data, we evaluate $\varepsilon_{\PEHE}$ on the validation dataset. For the Jobs experiments, we evaluate the policy risk $R_{\textup{POL}}$.
\begin{table}[H]
	\caption{Hyperparameter choices}
	\label{tab:hyperparameters}
	\begin{center}
		\begin{small}
			\resizebox{0.5\columnwidth}{!}{
				\begin{tabular}{ll}
					Hyperparameters & Value grid\\
					\hline
					Imbalance importance $\kappa$&$\{10^{k/2}\}_{k=-10}^{6}$\\
					Number of representations layers&$\{1,2,3,4,5\}$\\
					Dimensions of representations layers&$\{20,50,100,200\}$\\
					Batch size &$\{100,200,500\}$\\
					\hline
			\end{tabular}}
		\end{small}
	\end{center}
\end{table}
\subsection{DATASETS DETAILS}
The IHDP and Jobs datasets are public available already. For anonymmous purpose, we will provide the link to download those datasets upon being accepted. We also include the dataset in \textit{npz} files in the supplementary material. For the high-dimensional dataset, we provide a guidance of data generating process in the main text. The python script to generate this data is also supplied in the submitted code \textbf{high\_dim\_generating.py}. It will automatically generate the data for this experiments.
\subsection{COMPUTING INFRASTRUCTURE}
We run the code with environment \textbf{Tensorflow} \textbf{1.4.1} and \textbf{Numpy} \textbf{1.16.5} in \textbf{Python 2.7}. 
\subsection{PIPELINE FOR REPLICATIONS}
We provide guidance on running the experiments with the codebase provided in the supplementary material. The codebase is built on the one by \cite{shalit2017estimating}.
\begin{itemize}
	\item [1.]Create a directory for storing results. \textbf{mkdir results}
	\item [2.]Create a text file for all the hyperparameter you wish to sample with.
	\item [3.]Run \textbf{cfr\_param\_search.py dir n\_times}
	,where \textbf{dir} is the path for the configuration text, \textbf{n\_times} is the number of random sampling for the hyperparameters.
	\item [4.]Run \textbf{evaluate.py dir 1} to select the best model and evaluate performance on the testing data.
\end{itemize} 
We also provide an example configuration text file \textbf{IHDP\_1000\_EB.txt}
and a bash script \textbf{example\_ihdp\_1000.sh} for running the experiment in the supplementary material.
\vfill

\end{document}